\documentclass[letterpaper]{article} 
\usepackage{aaai2026}  
\usepackage{times}  
\usepackage{helvet}  
\usepackage{courier}  
\usepackage[hyphens]{url}  
\usepackage{graphicx} 
\usepackage{natbib}  
\usepackage{caption} 
\usepackage{algorithm}
\usepackage{algorithmic}

\usepackage{newfloat}
\usepackage{listings}
\usepackage{amsmath}
\usepackage{amssymb} 
\usepackage{amsthm}
\usepackage{multirow} 

\newtheorem{lemma}{Lemma}
\newtheorem{theorem}{Theorem}
\newtheorem{corollary}{Corollary}
\newtheorem{definition}{Definition}
\DeclareCaptionStyle{ruled}{labelfont=normalfont,labelsep=colon,strut=off} 
\floatstyle{ruled}
\newfloat{listing}{tb}{lst}{}
\floatname{listing}{Listing}
\title{DuoCast: Duo-Probabilistic Diffusion for Precipitation Nowcasting}
\author{
    Penghui Wen\textsuperscript{\rm 1},
    Mengwei He\textsuperscript{\rm 1},
    Patrick Filippi\textsuperscript{\rm 2},
    Na Zhao\textsuperscript{\rm 3}, \\
    Feng Zhang\textsuperscript{\rm 4},
    Thomas Francis Bishop\textsuperscript{\rm 2},
    Zhiyong Wang\textsuperscript{\rm 1},
    Kun Hu\textsuperscript{\rm 5,}\thanks{Corresponding author.}
}
\affiliations{

    \textsuperscript{\rm 1} School of Computer Science, The University of Sydney, Sydney, Australia\\

    \textsuperscript{\rm 2}	School of Life and Environmental Sciences, The University of Sydney, Sydney, Australia \\

    \textsuperscript{\rm 3} School of Mathematics, Shanghai University of Finance and Economics, Shanghai, China\\

    \textsuperscript{\rm 4}Department of Atmospheric and Oceanic Sciences, Institutes of Atmospheric Sciences, Fudan University, Shanghai, China\\
    \textsuperscript{\rm 5}School of Science, Edith Cowan University, Perth, Australia\\
    \{penghui.wen, mengwei.he, patrick.filippi\}@sydney.edu.au, zhaona@shufe.edu.cn, fengzhang@fudan.edu.cn, \{thomas.bishop, zhiyong.wang\}@sydney.edu.au, k.hu@ecu.edu.au
}

\usepackage{bibentry}

\begin{document}

\maketitle

\begin{abstract}
Accurate short-term precipitation forecasting is critical for weather-sensitive decision-making in agriculture, transportation, and disaster response. Existing deep learning approaches often struggle to balance global structural consistency with local detail preservation, especially under complex meteorological conditions. We propose \textit{DuoCast}, a dual-diffusion framework that decomposes precipitation forecasting into low- and high-frequency components modeled in orthogonal latent subspaces. We theoretically prove that this frequency decomposition reduces prediction error compared to conventional single branch U-Net diffusion models. In \textit{DuoCast}, the low-frequency model captures large-scale trends via convolutional encoders conditioned on weather front dynamics, while the high-frequency model refines fine-scale variability using a self-attention-based architecture. Experiments on four benchmark radar datasets show that \textit{DuoCast}  outperforms state-of-the-art baselines, achieving superior accuracy in both spatial detail and temporal evolution.

\end{abstract}

\begin{links}
    \link{Code}{https://github.com/ph-w2000/DuoCast}
    \link{Appendix}{https://arxiv.org/abs/2412.01091}
\end{links}

\section{Introduction}

Precipitation plays a vital role in weather systems, influencing temperature, humidity, and overall atmospheric conditions. Precipitation nowcasting, which predicts short-term rainfall using recent radar observations, is increasingly important for applications such as agriculture, transportation, and disaster response~\cite{zhang2023skilful}. Its importance was underscored during the July 2025 Texas flash floods: National Weather Service (NWS) observed localized totals over 5 inches and rainfall rates of 2–3 inches per hour at many locations, yet forecasts slightly mispredicted the heaviest cells, resulting in flash flooding in central Texas and trapping hundreds of people~\cite{cw3e2025-texas-floods-web}. Conventional approaches to nowcasting rely on numerical weather prediction (NWP), which simulates atmospheric dynamics by solving partial differential equations (PDEs) derived from physical laws~\cite{bauer2015quiet, lorenc1986analysis, alley2019advances}. However, the complexity of NWP renders them computationally expensive and inefficient for real-time forecasting~\cite{bi2023accurate}.

\begin{figure}[t]
  \centering
   \includegraphics[width=0.95\linewidth]{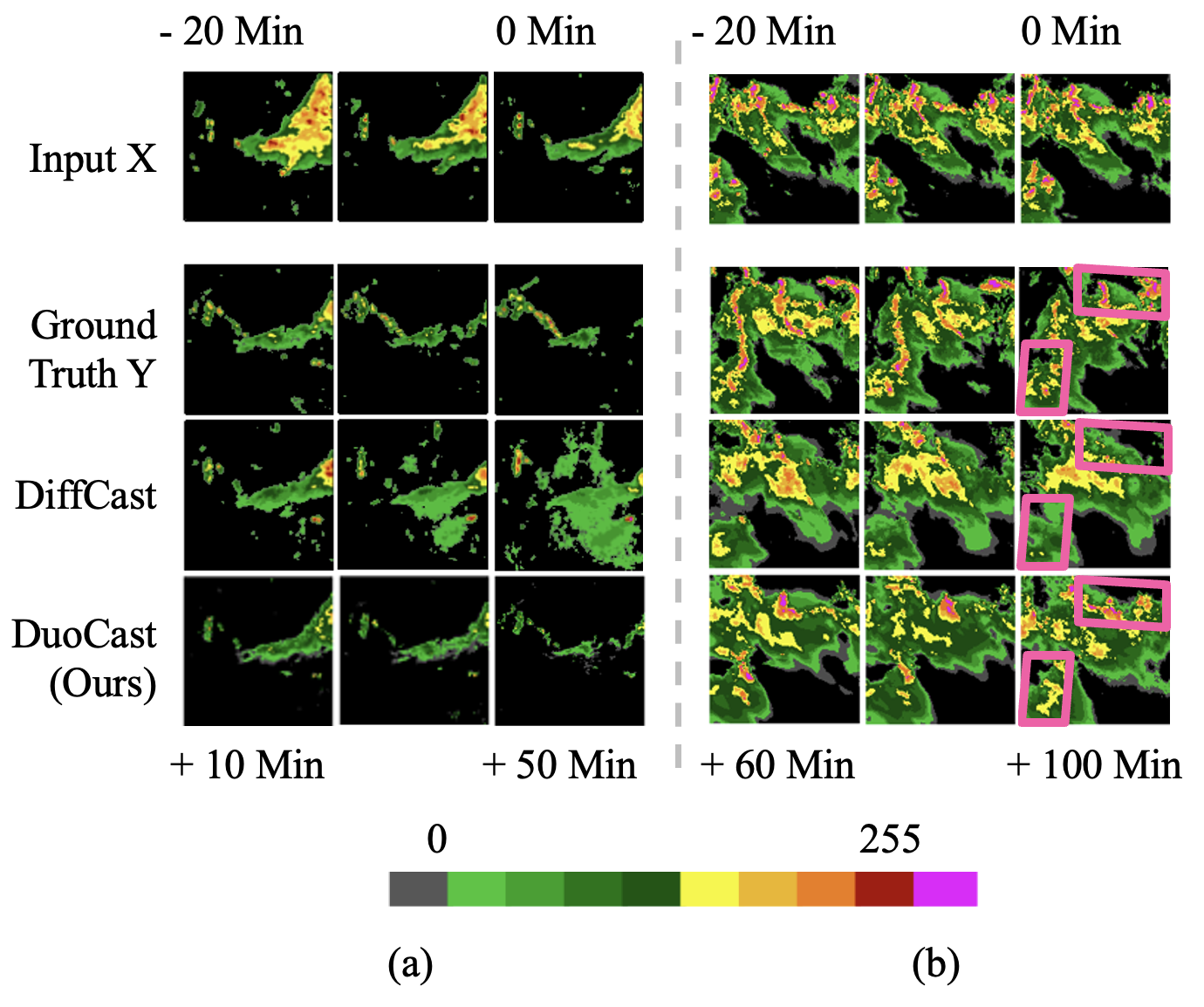}
   \caption{Challenges in precipitation nowcasting: a) precipitation with weather (warm) front patterns, and b) high-frequency features (micro-scale variability) within edge regions shown in pink parallelograms.}
   \label{fig:warm_micro}
\end{figure}

Recent advances in data-driven deep learning techniques have shown greater potential than traditional NWP methods in precipitation prediction by leveraging large datasets and avoiding the need to solve complex physical equations.
Deterministic models~\cite{shi2017deep, gao2022simvp, yan2024fourier, Lin_2025_CVPR} aim to capture the overall motion of precipitation systems by generating mean-value forecasts. However, they often produce blurry predictions and lack fine-grained spatial detail~\cite{ravuri2021skilful}. 
In contrast, probabilistic models~\cite{ravuri2021skilful, zhang2023skilful, gao2024prediff} introduce stochastic latent variables to better capture local-scale variability. While effective at modeling uncertainty, treating the entire system as fully stochastic can introduce excessive randomness and degrade forecasting precision~\cite{yu2024diffcast}.
To address this, hybrid approaches such as DiffCast~\cite{yu2024diffcast} combine global deterministic forecasting with local probabilistic refinement. Nevertheless, its CNN-based diffusion model struggles to preserve high-frequency information due to limited capacity~\cite{yang2023diffusion}, resulting in less spatially accurate precipitation intensity, as illustrated in Fig.~\ref{fig:warm_micro}(b).

Therefore, we introduce \textit{DuoCast}, a dual-diffusion framework for  precipitation nowcasting. \textit{DuoCast} decomposes the forecasting task into two frequency-specific subspaces, where separate stochastic diffusion models capture low- and high-frequency components. We theoretically show that frequency decomposition reduces prediction error compared to standard U-Net-based diffusion with formal proof. Specifically, the low-frequency diffusion model captures large-scale precipitation trends via convolutions, conditioned on weather front representations extracted from historical radar data. These features integrate spatial modeling of air masses and temporal modeling of front evolution, with intensity-aware enhancement for structural regions. The high-frequency model operates in latent space using a self-attention backbone to model fine-scale variability, conditioned on both historical inputs and the low-frequency forecast. Evaluated on four benchmark radar datasets, \textit{DuoCast} achieves state-of-the-art performance and yields substantial improvements in precipitation nowcasting accuracy.

Our work's key contributions are summarized as follows:
\begin{itemize}
\item We propose \textit{DuoCast}, a dual-diffusion framework that decomposes precipitation forecasting into low- and high-frequency subspaces for frequency-aware modeling.
\item We provide a theoretical analysis showing that frequency decomposition reduces prediction error compared to standard single-branch diffusion models.
\item Extensive experiments on four benchmark radar datasets demonstrate that \textit{DuoCast} achieves state-of-the-art performance in short-term precipitation nowcasting.
\end{itemize}

\section{Related Work}

\subsection{Precipitation Nowcasting}

Precipitation nowcasting has traditionally relied on numerical weather prediction (NWP), which solves atmospheric PDEs~\cite{ skamarock2008description}. Although effective, NWP is computationally intensive, often requiring hours on supercomputers~\cite{bi2023accurate}. 
Recent deep learning models~\cite{andrychowicz2023deep} offer competitive performance with significantly lower latency. They are generally categorized as either deterministic or probabilistic.

\noindent \textbf{Deterministic Precipitation Models} aim to predict the mean-based evolution of precipitation by modeling spatial and temporal dynamics. ConvLSTM~\cite{shi2015convolutional} captures these dynamics via convolutional and recurrent structures, while PhyDNet~\cite{guen2020disentangling} introduces a physics-guided decomposition into stochastic and deterministic motion. FACL~\cite{yan2024fourier} replaces standard L2 loss with a loss tailored for signal-based spatiotemporal modeling. AlphaPre~\cite{Lin_2025_CVPR} disentangles phase and amplitude to separately model motion and intensity.

\noindent \textbf{Probabilistic Precipitation Models} use latent variables to capture the inherent uncertainty of future weather, enabling finer modeling of micro-scale phenomena. Representative methods include DGMR~\cite{ravuri2021skilful}, which employs a generative adversarial network (GAN) with spatial and temporal discriminators, and NowcastNet~\cite{zhang2023skilful}, which integrates physical priors into GAN-based prediction. A diffusion model generates images by progressively denoising random noise into coherent visuals \cite{wen2024radio, lu2024autoregressive}. Building on this, Prediff~\cite{gao2024prediff} incorporates knowledge control to guide the sampling process. Although such approaches enhance forecast diversity, their fully stochastic nature can also introduce excessive variability, undermining predictive reliability.



\noindent \textbf{Hybrid Precipitation Models} combine deterministic backbones for global precipitation trends with probabilistic components to capture local variability, aiming to balance accuracy and diversity. For example, CasCast~\cite{gong2024cascast} and DiffCast~\cite{yu2024diffcast} integrate trajectory forecasting with stochastic refinement. However, their limited treatment of frequency-specific structures and micro- and macro-patterns may hinder long-range performance.

\subsection{Frequency Analysis and Modeling}
Frequency-domain analysis has shown strong effectiveness in image analysis and generation \cite{tan2024frequency, woo2022add, wen2023robust,chen2025f2net}, serving as a valuable complement to spatial-based approaches, which often struggle with artifacts. For instance, \cite{doloriel2024frequency} apply frequency-based masking to extract shared features, while \cite{zhou2024freqblender} categorize frequency components into semantic, structural, and noise levels to identify regions with unique spectral characteristics.

\begin{figure*}[t]
  \centering
   \includegraphics[width=\linewidth]{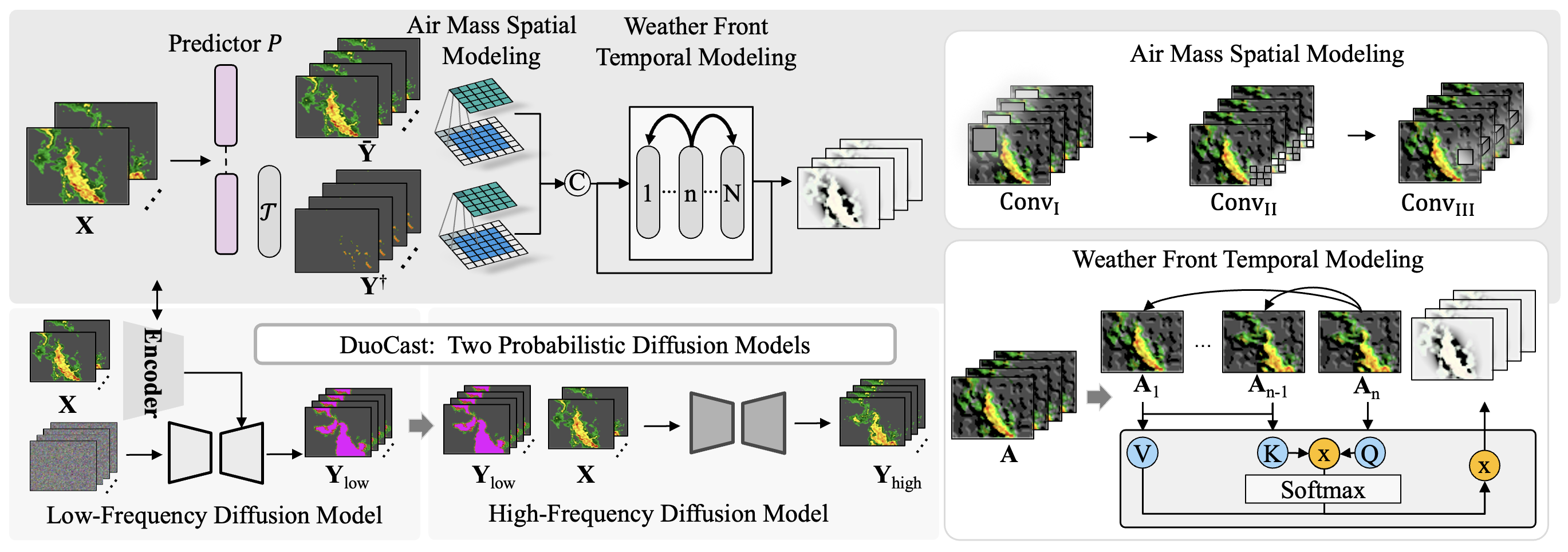}
   \caption{Overview of \textit{DuoCast}. A two-stage diffusion model leveraging historical weather fronts signals: (i) a low-frequency convolutional stage capturing front-guided precipitation trends, and (ii) a high-frequency self-attention stage refining micro-scale variability for pixel-level forecasts.
   }
   \label{fig:architecture}
\end{figure*}

\section{Methodology}

\subsection{Problem Formulation}
Precipitation nowcasting is formulated as a spatio-temporal forecasting task of single-channel radar echoes \cite{yu2024diffcast, gao2024prediff, gong2024cascast}. Given past echo intensity observations, denoted as $\mathbf{X} \in \mathbb{R}^{S\times H\times W}$ as the condition, short-term precipitation nowcasting models' objective is to model the conditional probabilistic $p(\mathbf{Y}|\mathbf{X})$ of the future following frames $\mathbf{Y} \in \mathbb{R}^{S \times H \times W}$, where $H$ and $W$ define the spatial resolution of each frame and $S$ is the number of temporal frames.
For notation simplicity, we denote $d=S\times H \times W$. To extend predictions over longer horizons, an autoregressive strategy is employed.

\subsection{Spectral Subspace Diffusion in DuoCast}
\noindent
To enhance precipitation estimation, \textit{DuoCast} employs two distinct stochastic diffusion processes to model low- and high-frequency components in dedicated subspaces.
The architecture overview is presented in Fig.~\ref{fig:architecture}.
Specifically, we design two complementary diffusion models:
low-frequency, which integrates weather front information to model low-frequency, large-scale precipitation patterns; and
high-frequency, which captures high-frequency, localized variability to refine fine-grained precipitation details.

Let $\mathcal{F}$ denote the Fourier transform on $L^2(\mathbb{R}^d)$, and write $\widehat{f} := \mathcal{F}f$ for the transform of $f$. Given a cutoff $\omega_c > 0$, define the low-frequency subspace $\mathcal{B}_{\omega_c} \subset L^2(\mathbb{R}^d)$ as
\begin{equation*}
\mathcal{B}_{\omega_c} := \left\{ f \in L^2(\mathbb{R}^d) : \mathrm{supp}(\widehat{f}) \subseteq \left\{ \xi \in \mathbb{R}^d : |\xi| \le \omega_c \right\} \right\},
\end{equation*}
and denote its orthogonal complement as $\mathcal{B}_{\omega_c}^\perp$. Define the corresponding orthogonal projections by
\begin{equation}
P_{\mathcal{B}_{\omega_c}} f := \mathcal{F}^{-1}(\chi_{\omega_c} \widehat{f}), 
P_{\mathcal{B}_{\omega_c}^\perp} f := f - P_{\mathcal{B}_{\omega_c}} f,
\end{equation}
where $\chi_{\omega_c}(\xi) := \mathbf{1}_{\{|\xi| \le \omega_c\}}(\xi)$ is the frequency-domain indicator function. These projections satisfy the standard orthogonal projection properties:
\begin{equation}
P_{\mathcal{B}_{\omega_c}}^2 = P_{\mathcal{B}_{\omega_c}}, 
P_{\mathcal{B}_{\omega_c}^\perp}^2 = P_{\mathcal{B}_{\omega_c}^\perp}, 
P_{\mathcal{B}_{\omega_c}} P_{\mathcal{B}_{\omega_c}^\perp} = 0.
\end{equation}

To this end, $\mathbf{Y}$ can be decomposed into $\mathbf{Y}_\text{low} = P_{\mathcal{B}_{\omega_c}}(\mathbf{Y})$ and $\mathbf{Y}_\text{high} = P_{\mathcal{B}_{\omega_c}^{\perp}}(\mathbf{Y})$. The two diffusion models estimate $\mathbf{Y}$ in two stages, for $\mathbf{Y}_\text{low}$ and $\mathbf{Y}_\text{high}$ to capture the low- and high-frequency components, respectively.

\paragraph{Low-Frequency Subspace Diffusion. }
Low-frequency components capture the overall scale and structure of precipitation systems, as macro trends. To estimate $\mathbf{Y_\text{low}}$ in the low-frequency subspace, a low-frequency diffusion network $\epsilon_{\theta_\text{low}}$ is employed, where $\theta_\text{low}$ indicates the parameters of the diffusion model.
For forward diffusion, we sample from $q(\mathbf{Y_\text{low}^\text{t}}|\mathbf{Y_\text{low}^\text{0}})$ in a closed form at an arbitrary timestep $t \in T$ as $q(\mathbf{Y_\text{low}^\text{t}} | \mathbf{Y_\text{low}^\text{t-1}})=\mathcal{N}(\mathbf{Y_\text{low}^\text{t}} ; \sqrt{1-\beta_{t}} \mathbf{Y_\text{low}^\text{t-1}}, \beta_{t} \mathbf{I})$. 
The reverse diffusion process is formulated as:
\begin{eqnarray}
p_{\theta_\text{low}}(\mathbf{Y}_\text{low} | \mathbf{X}).
\label{eq11}
\end{eqnarray}
To optimize ${\epsilon}_{\theta_\text{low}}$, an objective function is defined as follows:
\begin{eqnarray}
\mathcal{L}_\text{low}=\mathbb{E}_{\mathbf{Y}_\text{low}^0,\epsilon,t}\left\|\epsilon-{\epsilon}_{\theta_\text{low}}\left(\mathbf{Y}_{\text{low}}^t, t, \mathbf{X}\right)\right\|^2.
\label{eq21}
\end{eqnarray}

\paragraph{High-Frequency Subspace Diffusion.}
High-frequency components correspond to localized precipitation intensities, shown as micro-scale variations. 
To estimate $\mathbf{Y}_\text{high}$, we introduce a high-frequency diffusion model $\epsilon_{\theta_\text{high}}$, which follows a similar formulation as the low-frequency counterpart, as defined in Equations~\ref{eq11} and~\ref{eq21}.
To optimize ${\epsilon}_{\theta_\text{high}}$, an objective function is defined as follows:
\begin{eqnarray}
\mathcal{L}_\text{high}=\mathbb{E}_{\mathbf{Y}_\text{high}^0,\epsilon,t}\left\|\epsilon-{\epsilon}_{\theta_\text{high}}\left(\mathbf{Y}_{\text{high}}^t, t, \mathbf{X}\right)\right\|^2.
\label{eq2}
\end{eqnarray}

\subsection{Foundations of Spectral Subspace Diffusion}
\paragraph{Spectral Decay of Finite Support Convolutions.} While convolutional backbones like U-Net are commonly used in diffusion models, they are limited in generating high-frequency components—particularly crucial in precipitation nowcasting. We first illustrate the limitations of existing architectures in modeling high-frequency patterns, and then explain how a two-stage design operating in distinct spectral subspaces addresses this issue. 

\begin{lemma}[Polynomial Fourier decay of \textit{bounded variation} (BV) kernels]
Let \(k \in L^{1}(\mathbb{R}^{d})\) be a convolution kernel with compact support and
finite total variation, i.e.\ \(k \in BV(\mathbb{R}^{d})\).
Then there exists a constant \(C_{\mathrm{BV}}>0\) (depending only on \(\mathrm{TV}(k)\))
such that
\[
\bigl|\widehat{k}(\xi)\bigr| \;\le\;
{C_{\mathrm{BV}}}/({1+|\xi|}),
\qquad \forall\,\xi\in\mathbb{R}^{d}.
\]
\end{lemma}

The standard architectural constraints of CNNs (finite kernel size and finite‑precision weights) guarantee that the theoretic hypotheses of the lemma are met. Therefore the polynomial Fourier‑decay result is rigorously applicable to virtually all convolution kernels used in modern deep‑learning.
Suppose a depth-$L$ CNN (e.g., UNet or repeated UNet $\epsilon_\theta$ used in Diffusion) is formed by $L$ consecutive convolutional kernels $k_1, \dots, k_L$ with each $k_\ell \in BV(\mathbb{R}^d)$, 
and define the composite frequency response:
\begin{equation}
H_L(\xi) := \prod_{\ell=1}^{L} \widehat{k_\ell}(\xi).
\end{equation}
\begin{theorem}[Spectral envelope under bounded variation]\label{thm:bv_envelope}
Under the above assumptions, there exists $C > 0$ such that
\begin{equation}
|H_L(\xi)| \le C^L (1 + |\xi|)^{-L}, \qquad \forall\, \xi \in \mathbb{R}^{d}.
\end{equation}
\end{theorem}

\paragraph{Sharp Tail Bound.}
We now derive an \textit{input--dependent} high--frequency estimate that retains the true spectrum. 

\begin{theorem}[Sharp capacity bottleneck]\label{thm:bottleneck_sharp}
Let $\mathbf{Y}\in L^{2}(\mathbb{R}^d)$ be any target, $\omega_c=\Omega(\epsilon)$, $\mathbf{Y}_\text{low} = P_{\mathcal{B}_{\omega_c}}(\mathbf{Y})$ and $\mathbf{Y}_\text{high} = P_{\mathcal{B}_{\omega_c}^{\perp}}(\mathbf{Y}) = \mathbf{Y}-\mathbf{Y}_\text{low}$.
For every generation $g=\mathbf{Y}^t$ based on a CNN with input $f=\mathbf{Y}^0$,
\[
  \|\mathbf{Y}-\mathbf{Y}^t\|_{2}
  \;\ge\;
  \bigl[
      \|\mathbf{Y}_\text{high}\|_{2}
      -\sqrt{\varepsilon}\,\|\mathbf{Y}^0\|_{2}
  \bigr]_+.
\]
\end{theorem}

\begin{corollary}[Irreducible high‑frequency error]\label{cor:hf_gap}
If 
\(
  \|\mathbf{Y}_\text{high}\|_{2}>\sqrt{\varepsilon}\,\|\mathbf{Y}^0\|_{2},
\)
then every CNN-based estimation \(\mathbf{Y}^t\) obeys
\[
  \|\mathbf{Y}-\mathbf{Y}^t\|_{2}^{2}
  \;\ge\;
  \bigl(\|\mathbf{Y}_\text{high}\|_{2}-\sqrt{\varepsilon}\,\|\mathbf{Y}^0\|_{2}\bigr)^{2}
  \;>\;0,
\]
so the approximation error cannot be eliminated by any depth‑\(L\) CNN.
\end{corollary}

\paragraph{Two–Stage Diffusion Modeling. }
For any target $\mathbf{Y}=\mathbf{Y}_\text{low}+\mathbf{Y}_\text{high}$, where
$\mathbf{Y}_\text{low}\in\mathcal B_{\omega_c}$ and
$\mathbf{Y}_\text{high}\in\mathcal B_{\omega_c}^{\perp}$,
assume the first and second diffusion model generation
$g^{(1)}_{\theta}\colon\mathcal B_{\omega_c}\!\to\!\mathcal B_{\omega_c}$ and
$g^{(2)}_{\phi}\colon\mathcal B_{\omega_c}^{\perp}\!\to\!\mathcal B_{\omega_c}^{\perp}$, respectively.
Orthogonality yields:
\[
  \lVert \mathbf{Y}-g^{(1)}_{\theta}-g^{(2)}_{\phi} \rVert_{2}^{2}
  =\lVert \mathbf{Y}_\text{low}-g^{(1)}_{\theta}\rVert_{2}^{2}
  +\lVert \mathbf{Y}_\text{high}-g^{(2)}_{\phi}\rVert_{2}^{2}.
\]

\begin{corollary}[Two–stage universal approximation]
\label{cor:two_stage_exact}
Suppose the family $\mathcal G_{\theta}\subset\mathcal B_{\omega_c}$
is dense in $\mathcal B_{\omega_c}$, and
$\mathcal G_{\phi}\subset\mathcal B_{\omega_c}^{\perp}$
is dense in $\mathcal B_{\omega_c}^{\perp}$.
Then, for every $\mathbf{Y}=\mathbf{Y}_\text{low}+\mathbf{Y}_\text{high}\in L^{2}$,
\[
  \inf_{\theta\in\mathcal G_{\theta},\,\phi\in\mathcal G_{\phi}}
  \bigl\lVert \mathbf{Y}-g^{(1)}_{\theta}-g^{(2)}_{\phi}\bigr\rVert_{2}=0.
\]
\end{corollary}
We next investigate two (approximated) architectural designs for the low- and high-frequency diffusion models. 

\subsection{Low-Frequency Diffusion Model}
Weather fronts, which evolve gradually and span broad spatial regions, are key to shaping precipitation systems~\cite{catto2012relating}. For instance, warm fronts bring steady light rain, cold fronts induce abrupt heavy rainfall, and occluded fronts result in widespread precipitation. Inadequate modeling of them can lead to significant forecasting errors. As shown in Fig.~\ref{fig:warm_micro}(a), the context indicates a weakening warm front, yet DiffCast erroneously predicts intensifying rainfall. Therefore, the low-frequency diffusion captures these trends with convolutions to guide prediction, conditioned on front dynamics derived from underlying air mass interactions.

We first introduce a base encoder that extracts precipitation trends from the input sequence $\mathbf{X}$. It focuses on two key aspects: the spatial distribution of general precipitation and the identification of high-intensity precipitation regions.

Specifically, a convolutional predictor $P$ produces an initial forecast $\bar{\mathbf{Y}} = P(\mathbf{X})$, capturing the general spatial distribution of precipitation. To further isolate high-intensity regions that are indicative of strong air mass interactions, a thresholding operator $\mathcal{T}$ with cutoff $\theta_\text{int}$ is applied: $\mathbf{Y}^\dagger
 = \mathcal{T}(\bar{\mathbf{Y}}, \theta_\text{int})$.
These two outputs serve as proxies for identifying air mass locations and inferring evolving weather fronts. The predictor is trained via the following objective:
\begin{equation}
\mathcal{L}_{P} = \mathbb{E} \left\| \mathbf{Y} - \bar{\mathbf{Y}} \right\|^2.
\end{equation}

\paragraph{Air Mass Spatial Modeling. } 

To model the spatial organization of weather fronts, we introduce a convolutional module that learns the distribution and interactions of underlying air masses. As air mass boundaries often coincide with sharp gradients in precipitation~\cite{lagerquist2019deep}, this module extracts spatial cues relevant to front structure. Specifically, we apply three tailored convolutional layers to the preliminary forecasts $\bar{\mathbf{Y}}$ and $\mathbf{Y}^\dagger$ to capture local context, multi-scale structure, and temporal anchoring, forming a spatial representation $\mathbf{A}$.

\noindent\emph{1) Surrounding Mass Context - } $\text{Conv}_{\text{I}}(\cdot)$: 
We first apply large-kernel depth-wise convolutions to extract broad spatial contexts of air mass influence, 
capturing smooth but wide-spread structures such as pressure boundaries.

\noindent\emph{2) Occluded Front Dynamics - }  
$\text{Conv}_{\text{II}}(\cdot)$: 
To detect multi-scale interactions like occluded fronts, we apply dilated convolutions, which enhance sensitivity to overlapping cold and warm air regions at varying spatial scales.

\noindent\emph{3) Temporal Anchoring - }  
$\text{Conv}_{\text{III}}(\cdot)$: 
Lastly, we incorporate temporal locality via pixel-wise convolution across time,
which encodes the evolution of precipitation intensity at each spatial location.

The operations are applied separately to the general and high-intensity forecasts:
\begin{align}
\bar{\mathbf{A}} &= \text{Conv}_{\text{III}}(\text{Conv}_{\text{II}}(\text{Conv}_{\text{I}}(\bar{\mathbf{Y}}))), \notag \\
\mathbf{A}^\dagger &= \text{Conv}_{\text{III}}(\text{Conv}_{\text{II}}(\text{Conv}_{\text{I}}(\mathbf{Y}^\dagger))).
\end{align}
We concatenate the outputs to obtain the spatial air mass representation as front-aware spatial features:
\[
\mathbf{A} = \text{Concat}(\bar{\mathbf{A}}, \mathbf{A}^\dagger).
\]

\paragraph{Weather Front Temporal Modeling.}  
To capture the temporal evolution of weather fronts, we devise a lightweight convolution-based cross-attention mechanism that relates the current spatial features to both historical structure and recent trends. For each frame $\mathbf{A}_{n}$, we compute attention over two reference frames:  
(1) the leading frame $\mathbf{A}_{1}$, which serves as a static structural prior, and  
(2) the previous frame $\mathbf{A}_{n-1}$, which reflects recent changes in air mass interactions.
The attention query is derived from the current frame, while the key and value come from the concatenation of the leading and previous frames. Formally:
Cross-Attention($\mathbf{Q}=\mathbf{W}_{\mathbf{Q}} \circledast \mathbf{A}_{ n}$,$\mathbf{K}=\mathbf{W}_{\mathbf{K}} \circledast\left[\mathbf{A}_{1} \oplus \mathbf{A}_{n-1}\right]$,$\mathbf{V}=\mathbf{W}_{\mathbf{V}} \circledast\left[\mathbf{A}_{1} \oplus \mathbf{A}_{n-1}\right]$),
where $\circledast$ is the depth-wise convolution, $\oplus$ is concatenation along the channel dimension, and $\mathbf{W}_{\mathbf{Q}}$, $\mathbf{W}_{\mathbf{K}}$, and $\mathbf{W}_{\mathbf{V}}$ are learnable projection weights.

\subsection{High-Frequency Diffusion Model}

The {high-frequency} diffusion model $\epsilon_{\theta_2}$ captures high-frequency precipitation patterns using a fully self-attention-based architecture in latent space. By dynamically reweighting spatial features, it preserves fine-scale details more effectively than fixed convolutional filters~\cite{rombach2022high}. Complementing the coarse predictions from {low-frequency} diffusion, it enhances local variability and short-term dynamics, particularly at extended lead times.
The model is conditioned on two inputs: the structural prior $\hat{\mathbf{Y}}_\text{low}$ from {low-frequency} and the historical observations $\mathbf{X}$. These are encoded into a latent representation $\mathbf{Z}$ via a pretrained autoencoder~\cite{kingma2013auto}. The model finally generates a high-frequency correction $\hat{\mathbf{Y}}_\text{high}$. 

\subsection{Optimization}
We follow a standardized diffusion training process for both the {low-frequency} and {high-frequency} diffusion models:
\begin{eqnarray}
\mathcal{L}=\lambda_{1} \mathcal{L}_{P} + \lambda_{2}  \mathcal{L}_{\text{low}} + \lambda_{3}  \mathcal{L}_{\text{high}},
\label{loss_2}
\end{eqnarray}
where $\lambda_{1}$, $\lambda_{2}$ and $\lambda_{3}$ are hyperparameters.

\section{Experiments \& Discussions}

\subsection{Experimental Settings}

\subsubsection{Datasets}

To evaluate \textit{DuoCast}’s effectiveness in generating precise precipitation maps, we performed experiments using four radar echo datasets: SEVIR \cite{veillette2020sevir}, MeteoNet \cite{larvor2020meteo}, Shanghai\_Radar \cite{chen2020deep}, and CIKM\footnote{https://tianchi.aliyun.com/dataset/1085}.

\begin{table}[htbp]
\centering
\renewcommand{\arraystretch}{0.9}
\setlength{\tabcolsep}{3pt}
{%
\begin{tabular}{|l|ccccc|}
\hline
\multirow{2}{*}{Method} & \multicolumn{5}{c|}{SEVIR} \\
\cline{2-6}
 & \textuparrow CSI-M & \textuparrow CSI-181 & \textuparrow CSI-219 & \textuparrow HSS & \textuparrow SSIM  \\ 
\hline
DuoCast  & \textbf{0.338} & \textbf{0.182} & \textbf{0.107} & \textbf{0.432} & \textbf{0.683}\\
High-Freq & 0.310 & 0.158 & 0.050 & 0.401 & 0.611  \\
Low-Freq & 0.327 & 0.164 & 0.066 & 0.410 & 0.642  \\
\hline\hline
\multicolumn{6}{|c|}{Low-Frequency Backbone Exploration} \\
\hline
Attention & 0.310 & 0.143 & 0.043 & 0.393 & 0.608  \\
Conv & 0.327 & 0.164 & 0.066 & 0.410 & 0.642 \\
\hline
- $\text{Conv}_\text{I}$ & 0.316 & 0.155 & 0.051 & 0.405& 0.631\\
- $\text{Conv}_\text{II}$ & 0.317 & 0.159 & 0.048 & 0.400 & 0.632  \\
- $\text{Conv}_\text{III}$ & 0.320 & 0.156 & 0.050 & 0.404 & 0.622\\
- $\mathbf{Y^\dagger}$ & 0.316 & 0.146 & 0.048 & 0.400 & 0.619 \\ 
- Airmass & 0.309 & 0.143 & 0.051& 0.388 & 0.625  \\
- Front & 0.311 & 0.138 & 0.050 & 0.383 & 0.628  \\

\hline
\end{tabular}%
}
\caption{Ablation study on key components. \textit{Attention} and \textit{Conv} denote low-frequency models with self-attention and convolution backbones, respectively. \textit{Airmass} and \textit{Front} are air mass spatial and weather front temporal component. A dash (–) indicates component removal.}
\label{tab:ablation}
\end{table}

\subsubsection{Evaluation Metrics}
Following SOTA~\cite{yu2024diffcast, Lin_2025_CVPR, gong2024cascast}, we evaluate the nowcasting with the average Critical Success Index (CSI), Heidke Skill Score (HSS), Structural Similarity Index Measure (SSIM). The CSI, similar to the Intersection over Union (IoU), measures the degree of pixel-wise alignment between predictions and ground truth after thresholding them into binary (0/1) matrices, while HSS reflects improvement over random guessing. Following \cite{Lin_2025_CVPR}, we measure and average CSI (CSI-M) and CSI with two highest thresholds to measure the performance of high-intensity precipitation forecasting. SSIM assesses prediction quality.

\subsubsection{Implementation Details}
\textit{DuoCast} was trained with Adam optimizer (learning rate $1 \times 10^{-4}$) and 1,000 diffusion steps per low-/high-frequency branch. We used a two-stage schedule: first train the low-frequency model, then jointly train both. The learning rate was fixed across stages; to stabilize training we applied loss weights. In the first stage, the loss weight coefficients were $\lambda_1 = \lambda_2 = 0.5$. In the second stage, the loss weight were $\lambda_1 = \lambda_2 = 0.1$ and $\lambda_3 = 0.5$. The SimVP model is used as initial predictor $P$. The cutoff $\theta_\text{init}$ is set as the high-intensity precipitation threshold defined in the datasets.

\subsection{Compared with the State-of-the-Art}
To assess the performance of our \textit{DuoCast} framework in precipitation nowcasting, we compare it against a range of state-of-the-art methods, including probabilistic, deterministic, and hybrid approaches.

\begin{table*}[htbp]
\renewcommand{\arraystretch}{0.9}
\setlength{\tabcolsep}{4.5pt}
\centering
{
\begin{tabular}{|l|ccccc|ccccc|}
\hline
\multirow{2}{*}{Method} & \multicolumn{5}{c|}{SEVIR} & \multicolumn{5}{c|}{MeteoNet} \\
\cline{2-11}
 & \textuparrow CSI-M & \textuparrow CSI-181 & \textuparrow CSI-219 & \textuparrow HSS & \textuparrow SSIM & \textuparrow CSI-M & \textuparrow CSI-24 & \textuparrow CSI-32 & \textuparrow HSS & \textuparrow SSIM  \\
\hline
MAU (2021) & 0.3076&0.1071&0.0516&0.3863&0.6505&0.3233&0.2839&0.0997&0.4452&0.7897  \\
SimVP (2022)  & 0.3108&0.1106&0.0517&0.3924&0.6508&0.3351&0.3002&0.1130&0.4573&0.7804\\
FourCastNet (2022) &0.2686&0.0717&0.0339&0.3355&0.5976&0.3027&0.2533&0.1085&0.4216&0.6450 \\
Earthformer (2022)  & 0.2892&0.0844&0.0245&0.3665&0.6633&0.3205&0.2884&0.1237&0.4491&0.7772\\
PhyDNet (2020) &0.3017&0.1040&0.0278&0.3812&0.6532&0.3384&0.3194&0.1366&0.4673&0.7823 \\
EarthFarseer (2024) & 0.3004&0.0992&0.0413&0.3829&0.6327&0.3404&0.3170&0.1372&0.4726&0.7542 \\
NowcastNet (2023)  & 0.2791&0.0770&0.0351&0.3512&0.6839&0.3427&0.3206&0.1598&0.4751&0.7879\\
PreDiff (2024)  &0.2744&0.0627&0.0235&0.3592&0.5881&0.2944&0.2641&0.1202&0.4462&0.7085\\
DiffCast (2024) &0.3050&0.1300&0.0582&0.3996&0.6482&0.3512&0.3340&0.1808&0.4846&0.7887\\
CasCast (2024)  & 0.2878&0.0954&0.0312&0.3563&0.5680& 0.3317 & 0.3021 & 0.1351 & 0.4682 & 0.7622\\
FACL (2024)  & 0.3161 & \underline{0.1349} & \underline{0.0655} & 0.4033 & 0.5686& 0.3612 & 0.3341 & 0.1801 & 0.5006 & 0.7881 \\
AlphaPre (2025)  &\underline{0.3259}&0.1332&0.0545&\underline{0.4110}&\textbf{0.6884}&\underline{0.3824}&\underline{0.3633}&\underline{0.2002}&\underline{0.5164}&\underline{0.7968}\\

\textbf{Ours} & \textbf{0.3375}& \textbf{0.1818} & \textbf{0.1074} & \textbf{0.4318}& \underline{0.6827}& \textbf{0.3892}& \textbf{0.3841} & \textbf{0.2381} & \textbf{0.5297}& \textbf{0.7981} \\
\hline
\hline
\multirow{2}{*}{Method} & \multicolumn{5}{c|}{Shanghai Radar} & \multicolumn{5}{c|}{CIKM} \\
\cline{2-11}
 & \textuparrow CSI-M & \textuparrow CSI-35 & \textuparrow CSI-40 & \textuparrow HSS & \textuparrow SSIM & \textuparrow CSI-M & \textuparrow CSI-35 & \textuparrow CSI-40 & \textuparrow HSS & \textuparrow SSIM\\
\hline
MAU (2021)  & 0.3983&0.3621&0.2417&0.5346&0.7195&0.3039&0.2054&0.1241&0.3928&0.6325 \\
SimVP (2022)  &0.3850&0.3549&0.2382&0.5194&0.7795&0.3052&0.2044&0.1321&0.3955&0.6538 \\
FourCastNet (2022)  & 0.3571&0.3108&0.2073&0.4868&0.5598&0.2980&0.1849&0.1015&0.3801&0.4359 \\
Earthformer (2022) & 0.3503&0.3178&0.1872&0.4844&0.7298&0.3077&0.2039&0.1369&0.4001&0.6267 \\
PhyDNet (2020)  & 0.3654&0.3236&0.2176&0.4957&0.7751&0.3038&0.2052&0.1287&0.3931&0.6541\\
EarthFarseer (2024) & 0.3926&0.3608&0.2343&0.5330&0.5405&0.3000&0.2046&0.1259&0.3911&0.6373\\
NowcastNet (2023) & 0.3953&0.3608&0.2450&0.5334&\underline{0.7902}&0.2991&0.1940&0.1188&0.3865&0.6713 \\
PreDiff (2024) &0.3504&0.3026&0.1855&0.4991&0.6981&0.2842&0.1848&0.1002&0.3481&0.5582\\
DiffCast (2024)  & 0.4089&0.3740&0.2606&0.5476&0.7879&0.3159&0.2009&\underline{0.1457}&0.4085&0.6499\\
CasCast (2024)  & 0.3651 & 0.3291 & 0.2192 & 0.4971 & 0.7250 & 0.3021 & 0.1974 & 0.1142 & 0.3806 & 0.5241 \\
FACL (2024)  & 0.3940 & 0.3511 & 0.2355 & 0.5308 & 0.7601 & 0.3105 & 0.1993 & 0.1354 & 0.3996 & 0.6442 \\
AlphaPre (2025)  & \underline{0.4178}&\underline{0.3854}&\underline{0.2615}&\underline{0.5534}&\textbf{0.7951}&\textbf{0.3194}&\underline{0.2068}&0.1416&\underline{0.4137}&\underline{0.6568}\\

\textbf{Ours} & \textbf{0.4252}& \textbf{0.3933} & \textbf{0.2821} & \textbf{0.5643}& 0.7788  & \underline{0.3179}& \textbf{0.2275} & \textbf{0.1562} & \textbf{0.4217}& \textbf{0.6659} \\
\hline
\end{tabular}%
}
\caption{Quantitative comparison across different methods, datasets and evaluation metrics.}
\label{tab:sota}
\end{table*}

\subsubsection{Quantitative Analysis}

Based on the results in Table \ref{tab:sota}, we observe the following: i) Our \textit{DuoCast} framework achieves 2–4\% improvements in CSI and 2–5\% in HSS across four datasets. It also matches state-of-the-art performance in SSIM and MSE, showing its effectiveness in enhancing prediction accuracy. ii) The CSI at higher thresholds reflects \textit{DuoCast}’s capability to forecast high-intensity precipitation accurately. Furthermore, Fig. \ref{fig:csi_over_time} (a) illustrates the variation of CSI across different prediction time steps for recent methods, highlighting that our model consistently outperforms others across nearly all time steps.

\subsubsection{Qualitative Analysis}

\begin{figure}[t]
  \centering
   \includegraphics[width=0.99\linewidth]{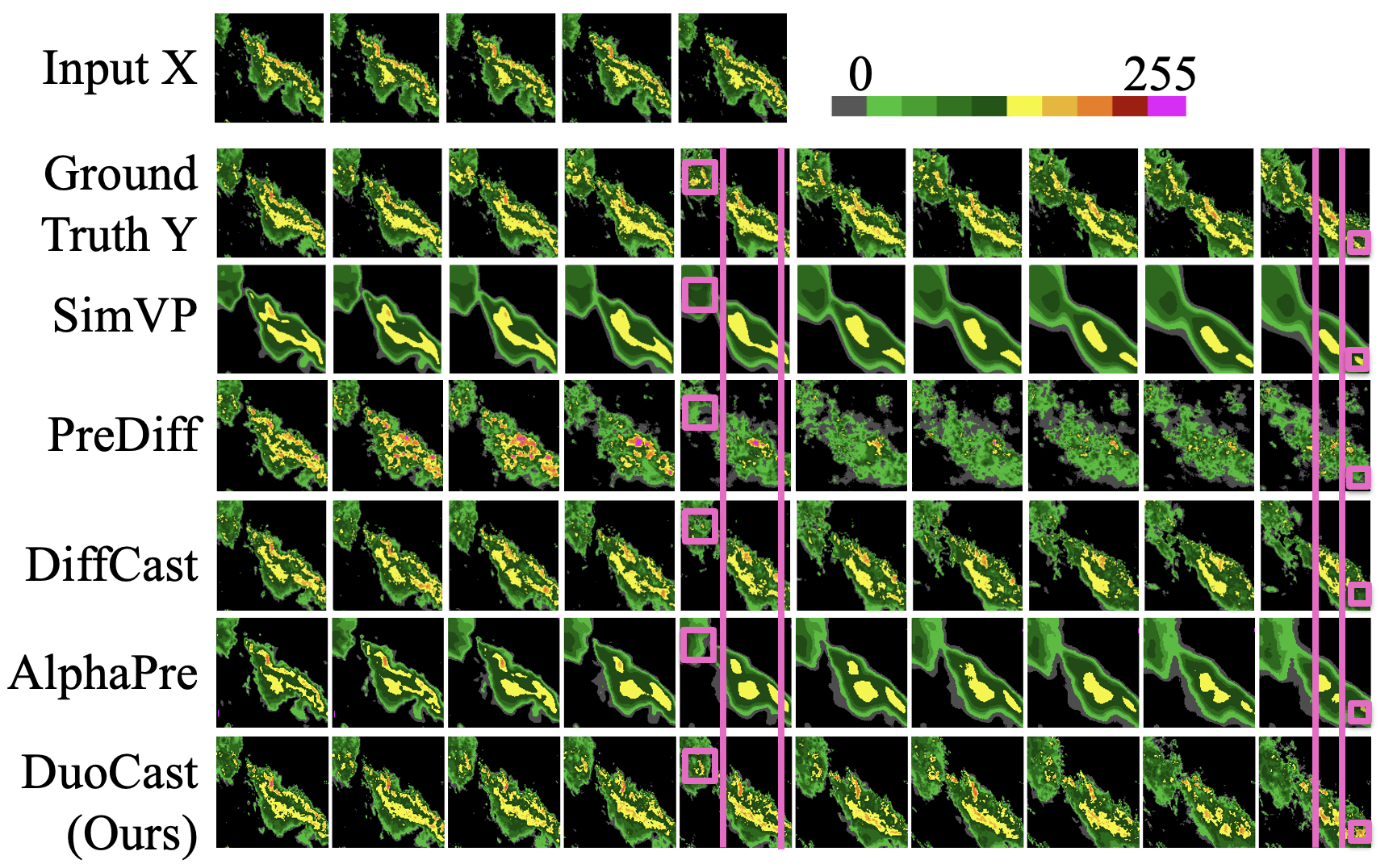}
   \caption{
   Qualitative comparison with SoTA on a SEVIR event. \textit{DuoCast} captures finer micro-scale details (pink box) and preserves more consistent evolution trends (pink line).
   }
   \label{fig:qualitative}
\end{figure}

Fig. \ref{fig:qualitative} presents a qualitative comparison of our model with various SoTA methods for a precipitation event. 
It can be observed that SimVP and AlphaPre struggle to capture micro-scale variability, resulting in fewer fine-grained details over longer lead times, as deterministic models cannot effectively represent stochastic behaviors.
While the probabilistic model Prediff capture finer micro-scale details, its predictions deviate significantly in trend, as probabilistic models often introduce excessive, uncontrollable randomness. In particular, Prediff forecasts a weakened precipitation band in the central region, resulting in an incorrect reduction in predicted rainfall.
Although DiffCast performs well in capturing both trend evolution and micro-scale variability in first a few frames, it cannot preserve high-frequency components, leading to overlooked regions on the map.  Specifically, at both 50 and 100 minutes prediction, it misses the rainfall in the top-left and bottom-right regions. 
In contrast, \textit{DuoCast} effectively captures both the overall trend of weather patterns and the finer micro-scale variability, with edge regions particularly catered.

\subsection{Ablation Study}

\subsubsection{Low-Frequency Diffusion Model.}
\begin{figure*}[t]
  \centering
   \includegraphics[width=0.96\linewidth]{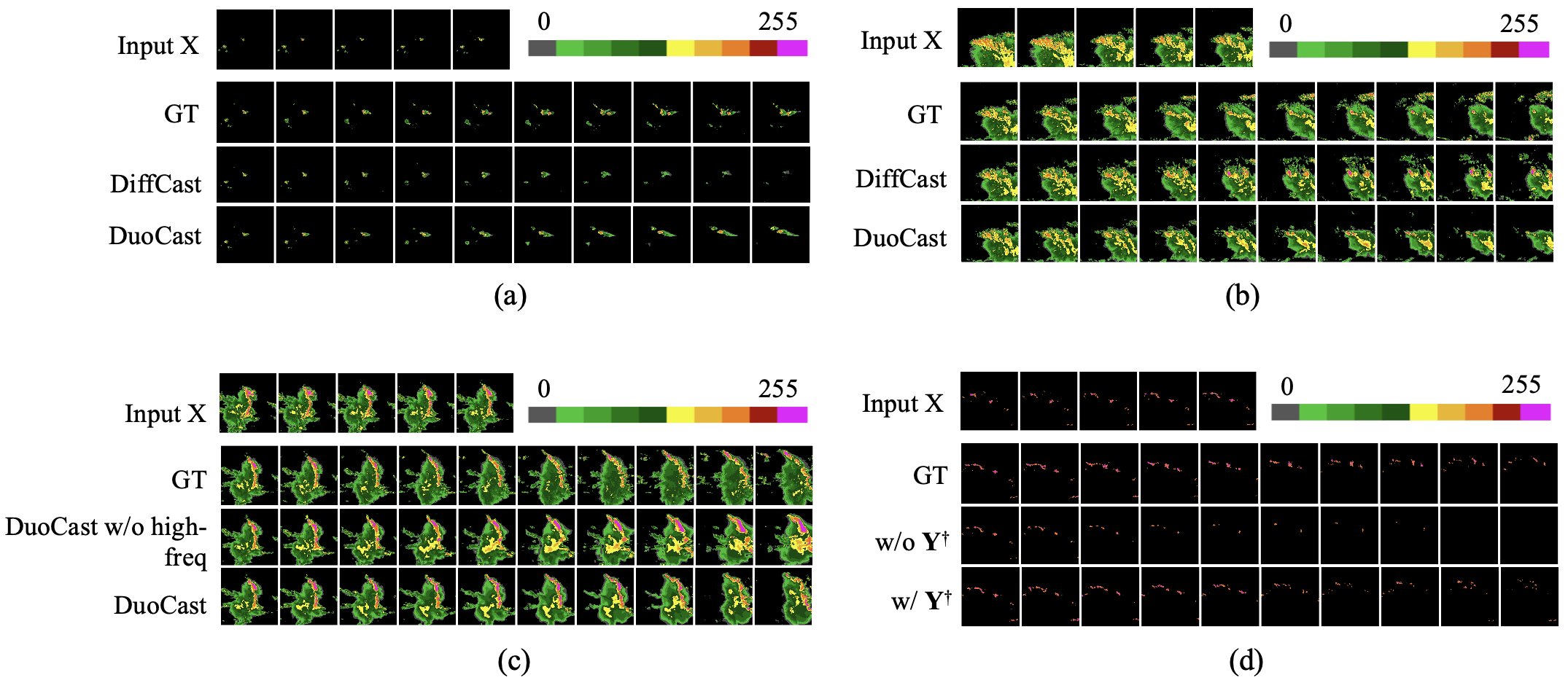}
   \caption{Ablation study with qualitative SEVIR examples. (a) and (b) demonstrate the effectiveness of \textit{DuoCast} in leveraging weather fronts, specifically cold and warm fronts, respectively. (c) highlights the capability of the high-frequency model in refining micro-scale variability. (d) showcases the impact of $\mathbf{Y^\dagger}$ for high-intensity regions prediction.}
   \label{fig:ablation}
\end{figure*}

To validate the effectiveness of the proposed mechanisms in \textit{DuoCast}, we compare our low-frequency model with the baseline DiffCast. As shown in Table~\ref{tab:ablation}, our model achieves higher CSI and HSS scores, demonstrating strong forecasting performance even with low-frequency modeling alone.
Qualitatively, Fig.~\ref{fig:ablation}(a) shows a cold front event where central high-precipitation regions expand over time. While the ground truth reflects this growth, DiffCast fails to capture the trend, predicting diminishing intensity. In contrast, our model accurately forecasts the expanding region, consistent with input observations.
Fig.~\ref{fig:ablation}(b) presents a warm front event where the central yellow precipitation area gradually dissipates. Again, DiffCast incorrectly predicts intensified rainfall, whereas our model captures the decreasing trend more faithfully.
These results show that the low-frequency model effectively captures large-scale precipitation by leveraging weather front dynamics. Additionally, we explore using a self-attention backbone, but it leads to a performance drop, because self-attention emphasizes inter-token fine-grained patterns making it less effective at preserving low-frequency structures.

\subsubsection{High-Frequency Diffusion Model. }

As shown in Table~\ref{tab:ablation}, removing the high-frequency diffusion model leads to a notable drop in CSI, HSS, and SSIM, highlighting its contribution to forecasting accuracy. Fig.~\ref{fig:ablation}(c) illustrates this effect: without the high-frequency component, the predicted precipitation band in the top-right region becomes increasingly blurred and blocky over time, especially between 80 and 100 minutes, missing fine-grained intensity variations. In contrast, incorporating the high-frequency model yields sharper predictions with more accurate spatial extent and intensity across multiple precipitation levels. These results demonstrate that the high-frequency diffusion process effectively refines coarse low-frequency outputs, enhancing micro-scale variability over extended lead times.

\subsubsection{Air Mass Modeling.} 
\begin{figure}[t]
  \centering
   \includegraphics[width=0.99\linewidth]{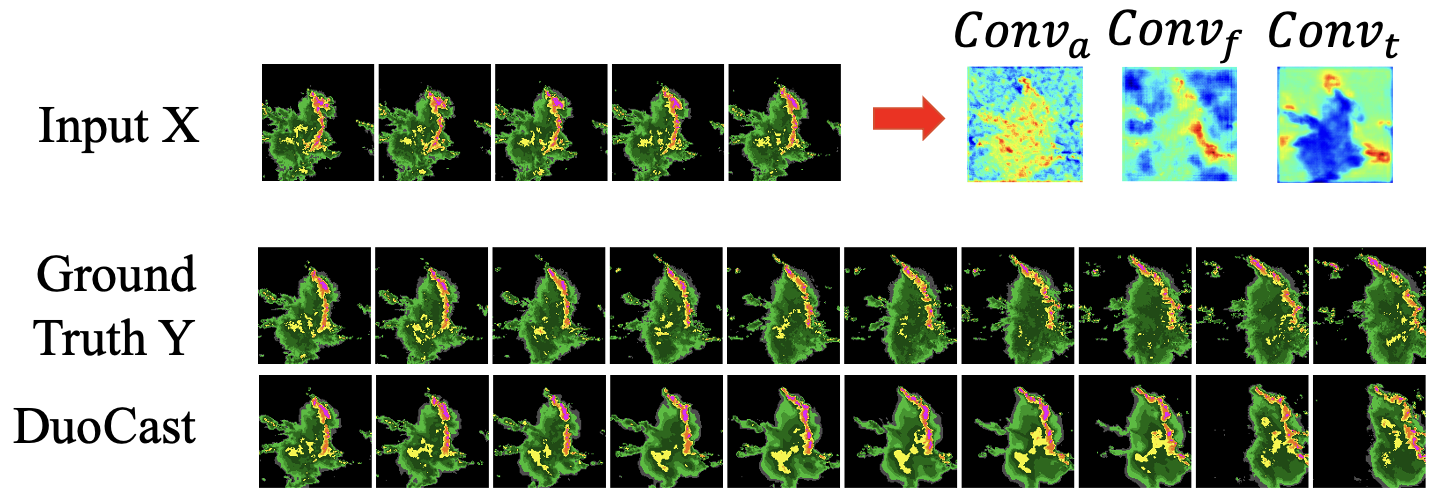}
   \caption{Visualization of air mass spatial modeling, highlighting captured weather fronts.
}
   \label{fig:ablation_air_conv}
\end{figure}

As shown in Table~\ref{tab:ablation}, removing air mass spatial modeling leads to a notable drop in CSI, particularly CSI-M, indicating its importance in capturing the spatial evolution of weather fronts. Additionally, we analyze the effectiveness of convolution in air mass spatial modeling. As shown in the Fig. \ref{fig:ablation_air_conv}, $\text{Conv}_\text{I}$, captures the spatial dynamics of surrounding air mass effects, highlighted by the heatmap. The frontal convolution, $\text{Conv}_\text{II}$, focuses on occluded fronts, which occur in critical transition areas where cold and warm air masses interact, as depicted by the figure's red band. The temporal convection convolution, $\text{Conv}_\text{III}$, captures pixel-level temporal convection, with the red areas indicating potential temporal motion. Quantitatively, as in Table \ref{tab:ablation}, every component in air mass spatial modeling contributes to superior performance.

\subsubsection{Weather Front Modeling.} 

Similarly, omitting weather front temporal modeling results in decreased CSI and HSS, reflecting its role in preserving temporal consistency. Additional ablation results for the temporal modeling component are provided in the supplementary material. These findings underscore the necessity of incorporating temporal front modeling for accurate precipitation forecasting.

\subsubsection{$\mathbf{Y^\dagger}$ for High-Intensity Precipitation}
\begin{figure}[t]
  \centering
   \includegraphics[width=\linewidth]{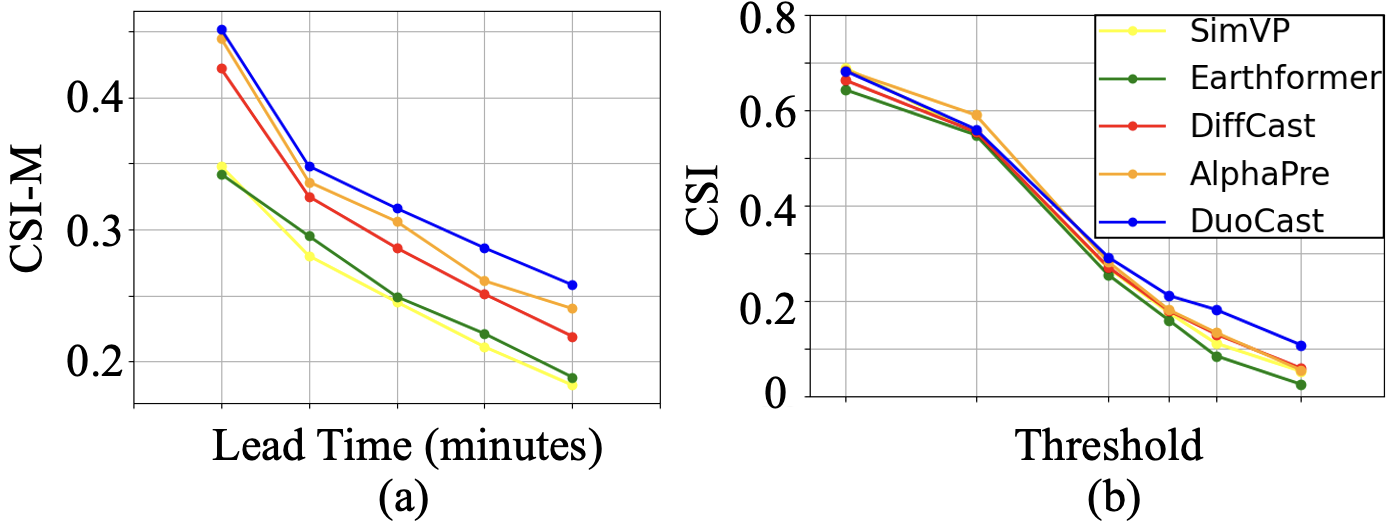}
   \caption{CSI across different (a) lead time steps and (b) intensity thresholds for different methods on SEVIR.}
   \label{fig:csi_over_time}
\end{figure}

To assess the role of $\mathbf{Y^\dagger}$ in modeling high-intensity precipitation, we conduct targeted experiments on SEVIR. As shown in Table~\ref{tab:ablation}, removing $\mathbf{Y^\dagger}$ results in substantial performance drops for severe rainfall events, with CSI-181 and CSI-219 decreasing by 10.8\% and 27.0\%, respectively. 
Figure~\ref{fig:ablation}(d) visualizes high-intensity events (values $\geq$ 181), where predictions with $\mathbf{Y^\dagger}$ align closely with the ground truth, consistently maintaining the precipitation band over time. In contrast, models without $\mathbf{Y^\dagger}$ increasingly miss key regions, and the predicted band degrades notably after 30 minutes. 
These results highlight the importance of $\mathbf{Y^\dagger}$ in learning temporal consistency and structure in severe precipitation. Figure~\ref{fig:csi_over_time}(b) shows consistently higher CSI across thresholds, confirming our model’s strength in forecasting high-intensity rainfall.

\section{Conclusion}

We introduce \textit{DuoCast}---a dual-probabilistic meteorology-aware model designed for preserving different frequency information. Low-frequency aims to capture the broader trend of weather patterns and high-frequency is to obtain high frequency micro-scale variability relevant to precipitation. Comprehensive experiments on four real-world datasets validate the effectiveness of the proposed framework. 

\section{Acknowledgments}
This work was supported by the Australian Research Council (ARC) Linkage Project \#LP230100294 and ECU Science Early Career and New Staff Grant Scheme.

\bibliography{aaai2026}

@article{bauer2015quiet,
  title={The quiet revolution of numerical weather prediction},
  author={Bauer, Peter and Thorpe, Alan and Brunet, Gilbert},
  journal={Nature},
  volume={525},
  number={7567},
  pages={47--55},
  year={2015},
  publisher={Nature Publishing Group UK London}
}

@article{shi2017deep,
  title={Deep learning for precipitation nowcasting: A benchmark and a new model},
  author={Shi, Xingjian and Gao, Zhihan and Lausen, Leonard and Wang, Hao and Yeung, Dit-Yan and Wong, Wai-kin and Woo, Wang-chun},
  journal={Advances in Neural Information Processing Systems},
  volume={30},
  year={2017}
}

@article{ravuri2021skilful,
  title={Skilful precipitation nowcasting using deep generative models of radar},
  author={Ravuri, Suman and Lenc, Karel and Willson, Matthew and Kangin, Dmitry and Lam, Remi and Mirowski, Piotr and Fitzsimons, Megan and Athanassiadou, Maria and Kashem, Sheleem and Madge, Sam and others},
  journal={Nature},
  volume={597},
  number={7878},
  pages={672--677},
  year={2021},
  publisher={Nature Publishing Group UK London}
}

@article{zhang2023skilful,
  title={Skilful nowcasting of extreme precipitation with NowcastNet},
  author={Zhang, Yuchen and Long, Mingsheng and Chen, Kaiyuan and Xing, Lanxiang and Jin, Ronghua and Jordan, Michael I and Wang, Jianmin},
  journal={Nature},
  volume={619},
  number={7970},
  pages={526--532},
  year={2023},
  publisher={Nature Publishing Group UK London}
}

@article{gong2024cascast,
  title={{CasCast}: Skillful High-resolution Precipitation Nowcasting via Cascaded Modelling},
  author={Gong, Junchao and Bai, Lei and Ye, Peng and Xu, Wanghan and Liu, Na and Dai, Jianhua and Yang, Xiaokang and Ouyang, Wanli},
  journal={International Conference on Machine Learning},
  year={2024}
}

@inproceedings{guen2020disentangling,
  title={Disentangling physical dynamics from unknown factors for unsupervised video prediction},
  author={Guen, Vincent Le and Thome, Nicolas},
  booktitle={IEEE/CVF Conference on Computer Vision and Pattern Recognition},
  pages={11474--11484},
  year={2020}
}

@article{veillette2020sevir,
  title={{SEVIR}: A storm event imagery dataset for deep learning applications in radar and satellite meteorology},
  author={Veillette, Mark and Samsi, Siddharth and Mattioli, Chris},
  journal={Advances in Neural Information Processing Systems},
  volume={33},
  pages={22009--22019},
  year={2020}
}

@article{chen2020deep,
  title={A deep learning-based methodology for precipitation nowcasting with radar},
  author={Chen, Lei and Cao, Yuan and Ma, Leiming and Zhang, Junping},
  journal={Earth and Space Science},
  volume={7},
  number={2},
  pages={e2019EA000812},
  year={2020},
  publisher={Wiley Online Library}
}

@article{catto2012relating,
  title={Relating global precipitation to atmospheric fronts},
  author={Catto, JL and Jakob, Christian and Berry, Gareth and Nicholls, Neville},
  journal={Geophysical Research Letters},
  volume={39},
  number={10},
  year={2012},
  publisher={Wiley Online Library}
}

@article{lorenc1986analysis,
  title={Analysis methods for numerical weather prediction},
  author={Lorenc, Andrew C},
  journal={Quarterly Journal of the Royal Meteorological Society},
  volume={112},
  number={474},
  pages={1177--1194},
  year={1986},
  publisher={Wiley Online Library}
}

@article{kingma2013auto,
  title={Auto-encoding variational bayes},
  author={Kingma, Diederik P and Welling, Max},
  journal={arXiv preprint arXiv:1312.6114},
  year={2013}
}

@inproceedings{yan2024fourier,
  title={Fourier Amplitude and Correlation Loss: Beyond
Using L2 Loss for Skillful Precipitation Nowcasting},
  author={Yan, Chiu-Wai and Foo, Shi Quan and Trinh, Van Hoan and Yeung, Dit-Yan and Wong, Ka-Hing and Wong, Wai-Kin},
  booktitle={NeurIPS},
  year={2024}
}

@inproceedings{yang2023diffusion,
  title={Diffusion probabilistic model made slim},
  author={Yang, Xingyi and Zhou, Daquan and Feng, Jiashi and Wang, Xinchao},
  booktitle={IEEE/CVF Conference on computer vision and pattern recognition},
  pages={22552--22562},
  year={2023}
}

@inproceedings{tan2024frequency,
  title={Frequency-aware deepfake detection: Improving generalizability through frequency space domain learning},
  author={Tan, Chuangchuang and Zhao, Yao and Wei, Shikui and Gu, Guanghua and Liu, Ping and Wei, Yunchao},
  booktitle={the AAAI Conference on Artificial Intelligence},
  volume={38},
  number={5},
  pages={5052--5060},
  year={2024}
}

@inproceedings{woo2022add,
  title={Add: Frequency attention and multi-view based knowledge distillation to detect low-quality compressed deepfake images},
  author={Woo, Simon and others},
  booktitle={the AAAI conference on artificial intelligence},
  volume={36},
  number={1},
  pages={122--130},
  year={2022}
}

@inproceedings{doloriel2024frequency,
  title={Frequency masking for universal deepfake detection},
  author={Doloriel, Chandler Timm and Cheung, Ngai-Man},
  booktitle={ICASSP 2024-2024 IEEE International Conference on Acoustics, Speech and Signal Processing (ICASSP)},
  pages={13466--13470},
  year={2024},
  organization={IEEE}
}

@article{zhou2024freqblender,
  title={Freqblender: Enhancing deepfake detection by blending frequency knowledge},
  author={Zhou, Jiaran and Li, Yuezun and Wu, Baoyuan and Li, Bin and Dong, Junyu and others},
  journal={Advances in Neural Information Processing Systems},
  volume={37},
  pages={44965--44988},
  year={2024}
}

@article{lagerquist2019deep,
  title={Deep learning for spatially explicit prediction of synoptic-scale fronts},
  author={Lagerquist, Ryan and Mcgovern, Amy and Gagne II, David John},
  journal={Weather and Forecasting},
  volume={34},
  number={4},
  pages={1137--1160},
  year={2019}
}

@inproceedings{rombach2022high,
  title={High-resolution image synthesis with latent diffusion models},
  author={Rombach, Robin and Blattmann, Andreas and Lorenz, Dominik and Esser, Patrick and Ommer, Bj{\"o}rn},
  booktitle={IEEE/CVF conference on computer vision and pattern recognition},
  pages={10684--10695},
  year={2022}
}

@inproceedings{wen2024radio,
  title={Radio Frequency Signal based Human Silhouette Segmentation: A Sequential Diffusion Approach},
  author={Wen, Penghui and Hu, Kun and Yua, Dong and Ning, Zhiyuan and Li, Changyang and Wang, Zhiyong},
  booktitle={2024 IEEE International Conference on Multimedia and Expo (ICME)},
  pages={1--6},
  year={2024},
  organization={IEEE}
}

@article{wen2023robust,
  title={Robust audio anti-spoofing with fusion-reconstruction learning on multi-order spectrograms},
  author={Wen, Penghui and Hu, Kun and Yue, Wenxi and Zhang, Sen and Zhou, Wanlei and Wang, Zhiyong},
  journal={Interspeech 2023},
  year={2023}
}

@article{chen2025f2net,
  title={F2Net: A Frequency-Fused Network for Ultra-High Resolution Remote Sensing Segmentation},
  author={Chen, Hengzhi and Feng, Liqian and Wu, Wenhua and Zhu, Xiaogang and Leo, Shawn and Hu, Kun},
  journal={arXiv preprint arXiv:2506.07847},
  year={2025}
}

@inproceedings{lu2024autoregressive,
  title={Autoregressive omni-aware outpainting for open-vocabulary 360-degree image generation},
  author={Lu, Zhuqiang and Hu, Kun and Wang, Chaoyue and Bai, Lei and Wang, Zhiyong},
  booktitle={Proceedings of the AAAI Conference on Artificial Intelligence},
  volume={38},
  number={13},
  pages={14211--14219},
  year={2024}
}

@article{bi2023accurate,
  title={Accurate medium-range global weather forecasting with 3D neural networks},
  author={Bi, Kaifeng and Xie, Lingxi and Zhang, Hengheng and Chen, Xin and Gu, Xiaotao and Tian, Qi},
  journal={Nature},
  volume={619},
  number={7970},
  pages={533--538},
  year={2023},
  publisher={Nature Publishing Group UK London}
}

@inproceedings{gao2022simvp,
  title={Simvp: Simpler yet better video prediction},
  author={Gao, Zhangyang and Tan, Cheng and Wu, Lirong and Li, Stan Z},
  booktitle={IEEE/CVF Conference on Computer Vision and Pattern Recognition},
  pages={3170--3180},
  year={2022}
}

@article{shi2015convolutional,
  title={Convolutional {LSTM} network: A machine learning approach for precipitation nowcasting},
  author={Shi, Xingjian and Chen, Zhourong and Wang, Hao and Yeung, Dit-Yan and Wong, Wai-Kin and Woo, Wang-chun},
  journal={Advances in Neural Information Processing Systems},
  volume={28},
  year={2015}
}

@article{gao2024prediff,
  title={Prediff: Precipitation nowcasting with latent diffusion models},
  author={Gao, Zhihan and Shi, Xingjian and Han, Boran and Wang, Hao and Jin, Xiaoyong and Maddix, Danielle and Zhu, Yi and Li, Mu and Wang, Yuyang Bernie},
  journal={Advances in Neural Information Processing Systems},
  volume={36},
  year={2024}
}

@inproceedings{yu2024diffcast,
  title={Diffcast: A unified framework via residual diffusion for precipitation nowcasting},
  author={Yu, Demin and Li, Xutao and Ye, Yunming and Zhang, Baoquan and Luo, Chuyao and Dai, Kuai and Wang, Rui and Chen, Xunlai},
  booktitle={IEEE/CVF Conference on Computer Vision and Pattern Recognition},
  pages={27758--27767},
  year={2024}
}

@article{skamarock2008description,
  title={A description of the advanced research WRF},
  author={Skamarock, William C and Klemp, JB and Dudhia, Jimy and Gill, David O and Barker, Dale M and Duda, Michael G and Huang, Xiang-Yu and Wang, Wei and Powers, Jordan G},
  journal={National Center for Atmospheric Research},
  volume={3},
  year={2008}
}

@inproceedings{larvor2020meteo,
  title={Meteonet: An open reference weather dataset for ai by m{\'e}t{\'e}o-france},
  author={Larvor, Gwenna{\"e}lle and Berthomier, Lea},
  booktitle={American Meteorological Society Meeting Abstracts},
  volume={101},
  pages={1--ii},
  year={2021}
}

@article{alley2019advances,
  title={Advances in weather prediction},
  author={Alley, Richard B and Emanuel, Kerry A and Zhang, Fuqing},
  journal={Science},
  volume={363},
  number={6425},
  pages={342--344},
  year={2019},
  publisher={American Association for the Advancement of Science}
}

@inproceedings{Lin_2025_CVPR,
    author    = {Lin, Kenghong and Zhang, Baoquan and Yu, Demin and Feng, Wenzhi and Chen, Shidong and Gao, Feifan and Li, Xutao and Ye, Yunming},
    title     = {AlphaPre: Amplitude-Phase Disentanglement Model for Precipitation Nowcasting},
    booktitle = {Computer Vision and Pattern Recognition Conference (CVPR)},
    month     = {June},
    year      = {2025},
    pages     = {17841-17850}
}

@article{andrychowicz2023deep,
  title={Deep learning for day forecasts from sparse observations},
  author={Andrychowicz, Marcin and Espeholt, Lasse and Li, Di and Merchant, Samier and Merose, Alexander and Zyda, Fred and Agrawal, Shreya and Kalchbrenner, Nal},
  journal={arXiv preprint arXiv:2306.06079},
  year={2023}
}

@misc{cw3e2025-texas-floods-web,
  author       = {Moore, Ben and Bartlett, Sam and Cordeira, Jay and Kalansky, Julie},
  title        = {CW3E Event Summary: Early July 2025 Central Texas Floods},
  howpublished = {\url{https://cw3e.ucsd.edu/cw3e-event-summary-early-july-2025-central-texas-floods/}},
  note         = {Center for Western Weather and Water Extremes (CW3E), Scripps Institution of Oceanography, UC San Diego. Accessed 2025-11-09},
  year         = {2025},
  month        = jul
}

@inproceedings{yang2025ffr,
  title={FFR: Frequency Feature Rectification for Weakly Supervised Semantic Segmentation},
  author={Yang, Ziqian and Zhao, Xinqiao and Wang, Xiaolei and Zhang, Quan and Xiao, Jimin},
  booktitle={Computer Vision and Pattern Recognition Conference},
  pages={30261--30270},
  year={2025}
}

\clearpage
\appendix
\twocolumn[
  \begin{@twocolumnfalse}
    \section*{\centering \Large Supplementary Material}
    \vspace{2em} 
  \end{@twocolumnfalse}
]

\section{Motivating Observation}
Precipitation nowcasting signals contain a high proportion of high-frequency components, which capture micro-scale intensity variations but are difficult for conventional convolution-based diffusion models to preserve. 
To address this, we propose DuoCast, a dual-diffusion framework that decomposes the forecasting task into separate low- and high-frequency components, each modeled within orthogonal latent subspaces for frequency-specific representation and generation. 
We quantify the frequency characteristics of precipitation data by applying the Discrete Fourier Transform (DFT) and following the approach of~\cite{yang2025ffr}, where high-frequency components are defined using a threshold parameter~$\mu$. 
As shown in Table~\ref{tab:proportions}, the SEVIR dataset contains nearly twice the proportion of high-frequency content compared to ImageNet when $\mu = 0.5$, highlighting the importance of modeling high-frequency signals in this domain.

\begin{table}[htbp]
\centering
{%
\begin{tabular}{|l|l|}
\hline  $\mu$ & $0.5$ \\
\hline ImageNet & 0.005561\\
\hline CIFAR100	& 0.005200 \\
\hline MSCOCO &	0.005760 \\
\hline \textbf{SEVIR} & 0.010059 \\
\hline
\end{tabular}%
}
\caption{The proportion of high-frequency components in different dataset.}
\label{tab:proportions}
\end{table}

\section{Formal Proofs}
This section presents the detailed theoretical analysis and complete proofs of the main results in the manuscript. 
\subsection{Foundations of Spectral Subspace Diffusion}
\paragraph{Spectral Decay of Finite Support Convolutions.} While convolutional backbones like U-Net are commonly used in diffusion models, they are limited in generating high-frequency components—particularly crucial in precipitation nowcasting. We first illustrate how these architectures fall short in modeling such fine-scale patterns.

\begin{lemma}[Polynomial Fourier decay of \textit{bounded variation} (BV) kernels]
Let \(k \in L^{1}(\mathbb{R}^{d})\) be a convolution kernel with compact support and
finite total variation, i.e.\ \(k \in BV(\mathbb{R}^{d})\).
Then there exists a constant \(C_{\mathrm{BV}}>0\) (depending only on \(\mathrm{TV}(k)\))
such that
\[
\bigl|\widehat{k}(\xi)\bigr| \;\le\;
\frac{C_{\mathrm{BV}}}{1+|\xi|},
\qquad \forall\,\xi\in\mathbb{R}^{d}.
\]
\end{lemma}
\begin{proof}
Apply integration by parts along each coordinate axis. 
\end{proof}
The standard architectural constraints of CNNs (finite kernel size and finite‑precision weights) guarantee that the theoretic hypotheses of the lemma are met. Therefore the polynomial Fourier‑decay result is rigorously applicable to virtually all convolution kernels used in modern deep‑learning.
Suppose a depth-$L$ CNN (e.g., UNet or repeated UNet $\epsilon_\theta$ used in Diffusion) is formed by $L$ consecutive convolutional kernels $k_1, \dots, k_L$ with each $k_\ell \in BV(\mathbb{R}^d)$, 
and define the composite frequency response:
\begin{equation}
H_L(\xi) := \prod_{\ell=1}^{L} \widehat{k_\ell}(\xi).
\end{equation}
\begin{theorem}[Spectral envelope under bounded variation]\label{thm:bv_envelope}
Under the above assumptions, there exists $C > 0$ such that
\begin{equation}
|H_L(\xi)| \le C^L (1 + |\xi|)^{-L}, \qquad \forall\, \xi \in \mathbb{R}^{d}.
\end{equation}
\end{theorem}

\paragraph{Sharp Tail Bound.}
We now derive an \textit{input--dependent} high--frequency estimate that retains the true spectrum of the driving signal $f$.

\begin{lemma}[Tail energy under BV envelope]\label{lem:tailA}
Let $f=\mathbf{Y^0_\text{low}}\in L^{2}(\mathbb R^{d})$ and denote $g := H_L * f$. For any radius $\Omega>0$,
\[
  \int_{|\xi|>\Omega} |\widehat g(\xi)|^{2}\,d\xi
  \le C^{2L}\!\int_{|\xi|>\Omega} (1+|\xi|)^{-2L}\,|\widehat f(\xi)|^{2}\,d\xi.
\]
If $L>d/2$, the integral on the right is finite.
\end{lemma}

\begin{proof}
Since $\widehat g = H_L\,\widehat f$ and $|H_L(\xi)|\le C^{L}(1+|\xi|)^{-L}$, the claim follows by direct substitution.
\end{proof}

\begin{corollary}[\(\varepsilon\)-tail radius]\label{cor:epsTailA}
Fix $\varepsilon\in(0,1)$. Any radius $\Omega_{\varepsilon}^{\sharp}(L)$ satisfying
\[
  C^{2L}\!\int_{|\xi|>\Omega_{\varepsilon}^{\sharp}(L)} (1+|\xi|)^{-2L}\,|\widehat f(\xi)|^{2}\,d\xi \le \varepsilon\,\|f\|_{2}^{2}
\]
ensures
\(
  \int_{|\xi|>\Omega_{\varepsilon}^{\sharp}(L)} |\widehat g(\xi)|^{2}\le \varepsilon\,\|f\|_{2}^{2}.
\)
\end{corollary}

\begin{definition}[Input–aware $\varepsilon$ cut‑off]
\label{def:eps_cutoff_sharp}
For $f\in L^{2}(\mathbb{R}^d)$ and $\varepsilon\in(0,1)$ define $\Omega_{\varepsilon}^{\sharp}(L)$ as:
\[
  \inf\Bigl\{
        \Omega>0:
        C^{2L}\!\!\int_{|\xi|>\Omega}
        (1+|\xi|)^{-2L}\,|\widehat f(\xi)|^{2}\,d\xi
        \le \varepsilon\,\|f\|_{2}^{2}
     \Bigr\}.
\]
By Lemma \ref{lem:tailA}, any feature
$g = H_L * f$ then satisfies
\(
  \displaystyle
  \int_{|\xi|>\Omega_{\varepsilon}^{\sharp}(L)}
      |\widehat g(\xi)|^{2}
  \le \varepsilon\,\|f\|_{2}^{2}.
\)
\end{definition}


\begin{theorem}[Sharp capacity bottleneck]\label{thm:bottleneck_sharp}
Let $\mathbf{Y}\in L^{2}(\mathbb{R}^d)$ be any target, $\omega_c=\Omega_{\varepsilon}^{\sharp}(L)$, $\mathbf{Y}_\text{low} = P_{\mathcal{B}_{\omega_c}}(\mathbf{Y})$ and $\mathbf{Y}_\text{high} = P_{\mathcal{B}_{\omega_c}^{\perp}}(\mathbf{Y}) = \mathbf{Y}-\mathbf{Y}_\text{low}$.
For every generation $g=\mathbf{Y}^t$ based on a CNN with input $f=\mathbf{Y}^0$,
\[
  \|\mathbf{Y}-\mathbf{Y}^t\|_{2}
  \;\ge\;
  \bigl[
      \|\mathbf{Y}_\text{High}\|_{2}
      -\sqrt{\varepsilon}\,\|\mathbf{Y}^0\|_{2}
  \bigr]_+.
\]
\end{theorem}
\begin{proof}
Write the orthogonal decompositions
$\mathbf{Y}^t = \mathbf{Y}^t_\text{Low}+\mathbf{Y}^t_\text{High}$ and 
$\mathbf{Y} = \mathbf{Y}_\text{Low}+\mathbf{Y}_\text{High}$ with
$\mathbf{Y}^t_\text{low} = P_{\mathcal{B}_{\omega_c}}(\mathbf{Y}^t)$ and $\mathbf{Y}^t_\text{high} = P_{\mathcal{B}_{\omega_c}^{\perp}}(\mathbf{Y}^t)$.
By Definition~\ref{def:eps_cutoff_sharp},
\[
  \|\mathbf{Y}^t_\text{High}\|_{2}^{2} \;=\;
  \!\int_{|\xi|>\omega_c}
     |\widehat g(\xi)|^{2}\,d\xi
  \;\le\; \varepsilon\,\|f\|_{2}^{2}.
\tag{1}
\]
\medskip
Parseval and projection orthogonality give
\[
  \|\mathbf{Y}-\mathbf{Y}^t\|_{2}^{2}
  \;=\;
  \|\mathbf{Y}_\text{High}-\mathbf{Y}^t_\text{High}\|_{2}^{2}
  +\|\mathbf{Y}_\text{Low}-\mathbf{Y}^t_\text{Low}\|_{2}^{2}.
\tag{2}
\]
Drop the first (non–negative) term:
\[
  \|\mathbf{Y}-\mathbf{Y}^t\|_{2}^{2}\;\ge\;\|\mathbf{Y}_\text{High}-\mathbf{Y}^t_\text{High}\|_{2}^{2}.
\tag{3}
\]
\medskip
Reverse triangle inequality in $L^{2}$:
\[
  \|\mathbf{Y}_\text{High}-\mathbf{Y}^t_\text{High}\|_{2}
  \;\ge\;
  \bigl|\|\mathbf{Y}_\text{High}\|_{2}-\|\mathbf{Y}_\text{High}^t\|_{2}\bigr|.
\tag{4}
\]
Combine with $\|g_{\mathrm H}\|_{2}\le\sqrt{\varepsilon}\|f\|_{2}$,
square both sides and obtain
\[
  \|\mathbf{Y}_\text{High}-\mathbf{Y}^t_\text{High}\|_{2}
  \;\ge\;
  \Bigl[\|\mathbf{Y}_\text{High}\|_{2}-\sqrt{\varepsilon}\,\|\mathbf{Y}^0\|_{2}\Bigr]_+.
\tag{5}
\]
\end{proof}

\begin{corollary}[Irreducible high‑frequency error]\label{cor:hf_gap}
If 
\(
  \|\mathbf{Y}_\text{High}\|_{2}>\sqrt{\varepsilon}\,\|\mathbf{Y}^0\|_{2},
\)
then every CNN-based estimation \(\mathbf{Y}^t\) obeys
\[
  \|\mathbf{Y}-\mathbf{Y}^t\|_{2}^{2}
  \;\ge\;
  \bigl(\|\mathbf{Y}_\text{High}\|_{2}-\sqrt{\varepsilon}\,\|\mathbf{Y}^0\|_{2}\bigr)^{2}
  \;>\;0,
\]
so the approximation error cannot be eliminated by any depth‑\(L\) CNN.
\end{corollary}

\paragraph{Two–Stage Diffusion Modeling. }
For any target $\mathbf{Y}=\mathbf{Y}_\text{Low}+\mathbf{Y}_\text{high}$, where
$\mathbf{Y}_\text{Low}\in\mathcal B_{\omega_c}$ and
$\mathbf{Y}_\text{High}\in\mathcal B_{\omega_c}^{\perp}$,
assume the first and second diffusion model generation
$g^{(1)}_{\theta}\colon\mathcal B_{\omega_c}\!\to\!\mathcal B_{\omega_c}$ and
$g^{(2)}_{\phi}\colon\mathcal B_{\omega_c}^{\perp}\!\to\!\mathcal B_{\omega_c}^{\perp}$, respecitively.
Orthogonality yields:
\[
  \lVert \mathbf{Y}-g^{(1)}_{\theta}-g^{(2)}_{\phi} \rVert_{2}^{2}
  =\lVert \mathbf{Y}_\text{Low}-g^{(1)}_{\theta}\rVert_{2}^{2}
  +\lVert \mathbf{Y}_\text{High}-g^{(2)}_{\phi}\rVert_{2}^{2}.
\]

\begin{corollary}[Two–stage universal approximation]
\label{cor:two_stage_exact}
Suppose the family $\mathcal G_{\theta}\subset\mathcal B_{\omega_c}$
is dense in $\mathcal B_{\omega_c}$, and
$\mathcal G_{\phi}\subset\mathcal B_{\omega_c}^{\perp}$
is dense in $\mathcal B_{\omega_c}^{\perp}$.
Then, for every $\mathbf{Y}=\mathbf{Y}_\text{Low}+\mathbf{Y}_\text{High}\in L^{2}$,
\[
  \inf_{\theta\in\mathcal G_{\theta},\,\phi\in\mathcal G_{\phi}}
  \bigl\lVert \mathbf{Y}-g^{(1)}_{\theta}-g^{(2)}_{\phi}\bigr\rVert_{2}=0.
\]
\end{corollary}

\begin{proof}
Fix an arbitrary $\delta>0$.  
By density of $\mathcal G_{\theta}$, choose
$\theta$ such that
$\lVert \mathbf{Y}_\text{Low}-g^{(1)}_{\theta}\rVert_{2}<\delta$.
Set $e_\text{Low}:=\mathbf{Y}_\text{Low}-g^{(1)}_{\theta}$.
Similarly, density of $\mathcal G_{\phi}$ provides
$\phi$ with
$\lVert \mathbf{Y}_\text{High}-g^{(2)}_{\phi}\rVert_{2}<\delta$;
define $e_\text{High}:=\mathbf{Y}_\text{High}-g^{(2)}_{\phi}$.
Because $e_{\mathrm L}\in\mathcal B_{\omega_c}$ and
$e_{\mathrm H}\in\mathcal B_{\omega_c}^{\perp}$ are orthogonal,
\[
  \lVert \mathbf{Y}-g^{(1)}_{\theta}-g^{(2)}_{\phi}\rVert_{2}^{2}
  =\lVert e_{\mathrm L}\rVert_{2}^{2}+\lVert e_{\mathrm H}\rVert_{2}^{2}
  <\delta^{2}+\delta^{2}=2\delta^{2}.
\]
Letting $\delta\to0$ proves the claim.
\end{proof}

\begin{table*}[htbp]
\centering
{%
\begin{tabular}{|p{4cm}|cccccc|}
\hline
\multirow{2}{*}{Method} & \multicolumn{6}{c|}{SEVIR} \\ 
\cline{2-7}
 & \textuparrow CSI & \textuparrow CSI-181 & \textuparrow CSI-219 & \textuparrow HSS & \textuparrow SSIM & \textdownarrow MSE \\ 
\hline
 $\lambda_{1}$ = 0.5 , $\lambda_{2}$ = 0.1 , $\lambda_{3}$ = 0.1  & 0.3255 & 0.1712 & 0.0963 & 0.4263 & 0.6753 & 480.83 \\ 
$\lambda_{1}$ = 0.1 , $\lambda_{2}$ = 0.5 , $\lambda_{3}$ = 0.1  & 0.3247 & 0.1754 & 0.0925 & 0.4048 & 0.6682 & 490.13 \\ 
$\lambda_{1}$ = 0.1 , $\lambda_{2}$ = 0.1 , $\lambda_{3}$ = 0.5  & \textbf{0.3375} & \textbf{0.1818} & \textbf{0.1074} & \textbf{0.4318} & \textbf{0.6827} & \textbf{463.07} \\ 
\hline
\end{tabular}%
}
\caption{Effect of loss weight hyper-parameters in Eq.~(10) on SEVIR for the DuoCast model.}
\label{tab:loss_ablation}
\end{table*}

\begin{table}[h!tbp]
\centering
\setlength{\tabcolsep}{0.7pt}
{%
\begin{tabular}{|c|cccccc|}
\hline
\multirow{2}{*}{Method} & \multicolumn{6}{c|}{SEVIR} \\ 
\cline{2-7}
 & \textuparrow CSI-M & \textuparrow CSI-181 & \textuparrow CSI-219 & \textuparrow HSS & \textuparrow SSIM & \textdownarrow MSE \\ 
\hline
w/o $\mathbf{A}_{1}$ & 0.320 & 0.157 & 0.060 & 0.398 & 0.633 & 465.4 \\ 
w/ $\mathbf{A}_{1}$ & \textbf{0.327} & \textbf{0.164} & \textbf{0.066} & \textbf{0.410} & \textbf{0.642} & \textbf{462.0} \\ 
\hline
\end{tabular}%
}
\caption{Ablation study showing the effectiveness of $\mathbf{A}_{1}$.}
\label{tab:a1_ablation}
\end{table}

\section{Datasets Details}
The effectiveness of DuoCast was evaluated on four widely used radar echo datasets: SEVIR, MeteoNet, Shanghai\_Radar and CIKM\footnote{https://tianchi.aliyun.com/dataset/1085}.

\noindent\textbf{SEVIR} captures both storm and random events in the United States from 2017 to 2019. It contains 20,393 sequences of radar frames representing weather events, each covering a 4-hour period with a spatial coverage of 384 km$\times$384 km. Every pixel corresponds to an area of 1 km$\times$1 km, and the data has a 5-minute temporal resolution. 

\noindent\textbf{MeteoNet} contains rain radar data from the northwest and southeast regions of France, covering the period from 2016 to 2018. The radar data in MeteoNet has a spatial resolution of $0.01^{\circ}$, with observations recorded at a 6-minute temporal resolution. Following SoTA methods, we used radar observations specifically from the northwest region of France.

\noindent\textbf{Shanghai\_Radar} consists of continuous radar echo frames generated by volume scans at an approximately 6-minute temporal resolution, collected from October 2015 to July 2018 in Pudong, Shanghai. Each radar echo map covers an area of 501 km$\times$501 km.

\noindent\textbf{CIKM} is from the CIKM AnalytiCup 2017 Competition, as a radar dataset that records precipitation samples over an 101 km$\times$101 km area in Guangdong, China. Each sequence includes 15 radar echo maps as a sample, with a 6-minute temporal resolution.

We follow the experimental protocols of state-of-the-art methods, such as DiffCast and AlphaPre, across all datasets, maintaining the original temporal resolution while reducing the spatial dimensions to $128\times128$. 
Additionally, since our primary focus is on modeling precipitation events, we segment continuous sequences into multiple events for the SEVIR, MeteoNet, Shanghai\_Radar, and CIKM datasets, exactly based on the procedure used in prior SOTA method.
For all datasets except CIKM, we predict 20 future frames given 5 past frames ($5 \rightarrow 20$, i.e., $L_\text{in} = 5$, $L_\text{out} = 20$). For CIKM, due to its shorter sequence length, we predict only 10 future frames given 5 past frames ($5 \rightarrow 10$, i.e., $L_\text{in} = 5$, $L_\text{out} = 10$).

\begin{table}[htbp]
\centering
\resizebox{\columnwidth}{!}{
\begin{tabular}{ccccccc}
\hline Dataset & $N_{t r}$ & $N_{v a}$ & $N_{t e}$ & $(C, H, W)$ & $T_i$ & $T_o$ \\
\hline SEVIR & 23808 & 6016 & 8100 & $(1,128,128)$ & 5 & 20 \\
MeteoNet & 6308 & 1310 & 1310 & $(1,128,128)$ & 5 & 20 \\
Shanghai & 1534 & 526 & 526 & $(1,128,128)$ & 5 & 20 \\
CIKM & 8000 & 2000 & 4000 & $(1,128,128)$ & 5 & 10 \\
\hline
\end{tabular}
}
\caption{Dataset statistics. $N_{t r}$, $N_{v a}$ and $N_{t e}$ represent the number of samples in the training, validation, and test sets, respectively.}
\label{tab:a1_ablation}
\end{table}

\section{Implementation Details}

We trained the \textit{DuoCast} framework using the Adam optimizer with a learning rate of 0.0001. In line with standard practices for diffusion models, we used 1,000 diffusion steps for both the low- and high-frequency branches. The batch sizes were set to 4 for the low-frequency model and 3 for the high-frequency model. We adopt a two-stage training strategy. In the first stage, only the low-frequency model is trained. In the second stage, both the low- and high-frequency models are trained jointly. The learning rate is kept the same in both stages. However, we apply loss weighting via $\lambda$ to stabilize training. In the first stage, the loss weight coefficients were empirically set to $\lambda_1 = \lambda_2 = 0.5$. In the second stage, the loss weight coefficients were empirically set to $\lambda_1 = \lambda_2 = 0.1$ and $\lambda_3 = 0.5$ without further tuning. The SimVP model is used as initial predictor $P$. The cutoff $\theta_\text{init}$ is set as the high-intensity precipitation threshold defined in the datasets SEVIR dataset $\theta_\text{init}$=181, MeteoNet dataset $\theta_\text{init}$=24, Shanghai and CIKM $\theta_\text{init}$=35.

\section{Additional Analysis}
In this section, we provide an extra analysis on the design of framework loss, the computational complexity and the effectiveness of our model design.

\noindent\textbf{Hyper-Parameters for Loss Weights.} 
Starting from an empirical setting ($\lambda_{1} = \lambda_{2} = 0.1$ and $\lambda_{3} = 0.5$), 
We conducted sensitivity experiments (Table~\ref{tab:loss_ablation}) to assess the impact of different weight configurations. The results indicate that the model performs robustly across a range of reasonable settings, with the combination $(0.1, 0.1, 0.5)$ achieving the best overall performance across all metrics. In contrast, the $(0.5, 0.1, 0.1)$ setting assigns equal emphasis to the low- and high-frequency models, causing the initially well-trained low-frequency model to become significantly biased as it adjusts to fit the high-frequency component during subsequent training. Similarly, the $(0.1, 0.5, 0.1)$ setting gives excessive weight to the low-frequency model, leading to over-optimization of the low-frequency component while leaving the high-frequency model under-trained.

\begin{figure}[t]
  \centering
   \includegraphics[width=\linewidth]{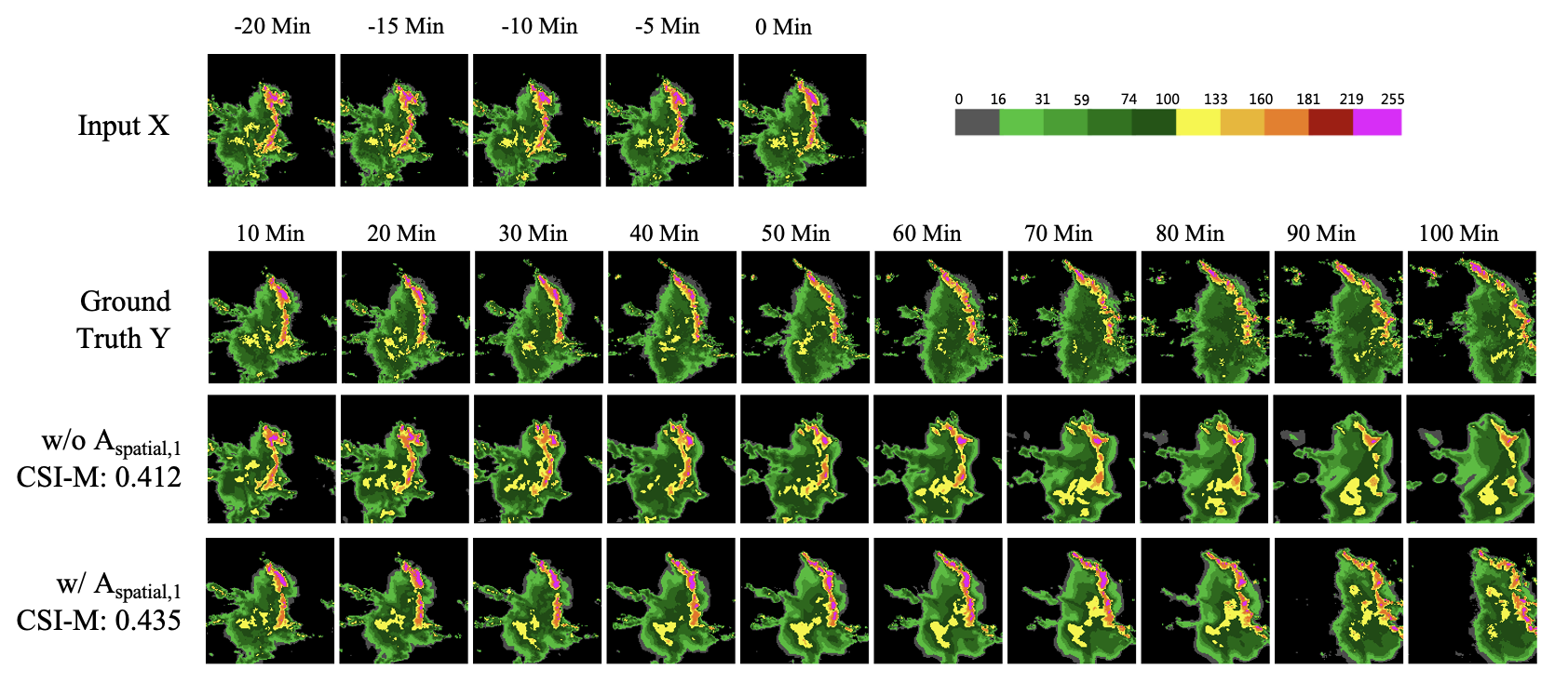}
   \caption{Qualitative comparison between w/o and w/ $\mathbf{A}_{1}$ on SEVIR.}
   \label{fig:ablation_a1}
\end{figure}

\noindent\textbf{Model Complexity.} 
In Table~\ref{tab:complexities}, we compare the number of parameters and TFLOPs of our method and existing baselines, evaluated on a single NVIDIA RTX 4090 GPU using the SEVIR dataset. DuoCast has more parameters than DiffCast but requires fewer TFLOPs, indicating improved computational efficiency. Compared to AlphaPre, DuoCast has a similar parameter count but incurs higher computational cost in terms of TFLOPs. However, given the significant advances in GPU hardware—such as the NVIDIA RTX 4090, which supports up to 82.6 TFLOPs (FP32) and 330.4 TFLOPs (FP16), DuoCast’s demand of 42.16 TFLOPs remains well within the capabilities of modern hardware. Considering the importance of inference speed in precipitation nowcasting, DuoCast achieves a practical balance between performance and efficiency. 

Additionally, DuoCast is efficient, with inference taking 0.56 seconds per sample (1 sample = 20 frames prediction; 100-minute horizon for SEVIR) on an NVIDIA RTX 4090, while the time of DiffCast is 3.67 seconds. This latency is suitable for running precipitation nowcasting smoothly on consumer GPUs.

\begin{table}[htbp]
\centering
{
\begin{tabular}{|l|ccc|}
\hline
\multirow{2}{*}{Method} & \multicolumn{3}{c|}{Complexity}\\
\cline{2-4}
 & \# Param & TFLOPs & Year \\ 
\hline
MCVD  & 105.91M & 4.69 & 2022\\
Prediff &  135.24M & 2.80 & 2023\\
STRPM & 439.63 M & 0.28 & 2022\\
SimVP  & 44.25M &0.05 & 2022\\
Earthformer & 34.61M & 0.04 &2022\\
MAU & 20.13M & 0.09 &2021\\
ConvGRU  &18.21M & 0.03 & 2017\\
PhyDnet & 11.80M & 0.08 &2020 \\
DiffCast  & 58.33M & 72.49 &2024\\
FACL  & 14.38M & 0.02 &2024\\ 
AlphaPre  & 89.03M & 1.56 &2025\\ \hline
DuoCast & 94.18M & 0.30 &2025\\
\hline
\end{tabular}%
}
\caption{Analysis of model complexity with SOTA methods.}
\label{tab:complexities}
\end{table}


\noindent\textbf{Effectiveness of $\mathbf{A}_{1}$ for Temporal Modeling.}
The gap between $\mathbf{A}_{1}$ and $\mathbf{A}_{n-1}$ captures essential temporal dynamics. Excluding $\mathbf{A}_1$ removes the model’s global temporal anchor, making it difficult to maintain coherent long-range structure. As a result, low-frequency components begin to drift over time, leading to distorted large-scale patterns and reduced forecast stability. In Fig. \ref{fig:ablation_a1}, without $\mathbf{A}_{1}$, the prediction aligns closely with the previous frame but fails to identify the change trend. For instance, in the 100-minute frame, though the high-intensity precipitation band aligns well with the preceding frame, its shape becomes distorted and diverges significantly from that in the first frame. The quantitative results in Table \ref{tab:a1_ablation} further support this observation.

\noindent\textbf{The Statistical Significance of Improvement.}
To assess statistical significance, we conducted paired sample t-tests between DuoCast and AlphaPre on the SEVIR dataset (CSI) across 8100 samples. Table \ref{tab:t-test} shows significant improvements. This confirms that DuoCast’s improvements are statistically significant.

\begin{table}[htbp]
\centering
{%
\begin{tabular}{|l|l|l|}
\hline  CSI-Threshold & p-value & DuoCast $>$ AlphaPre \\
\hline M & 0.035 & Significant difference\\
\hline 181	&  $<$0.001 & Significant difference\\
\hline 219 &0.001 & Significant difference\\
\hline
\end{tabular}%
}
\caption{The paired sample t-tests between DuoCast and AlphaPre on the SEVIR dataset (CSI).}
\label{tab:t-test}
\end{table}

\section{More Qualitative Results}
\begin{figure*}[htbp]
  \centering
   \includegraphics[width=\linewidth,keepaspectratio]{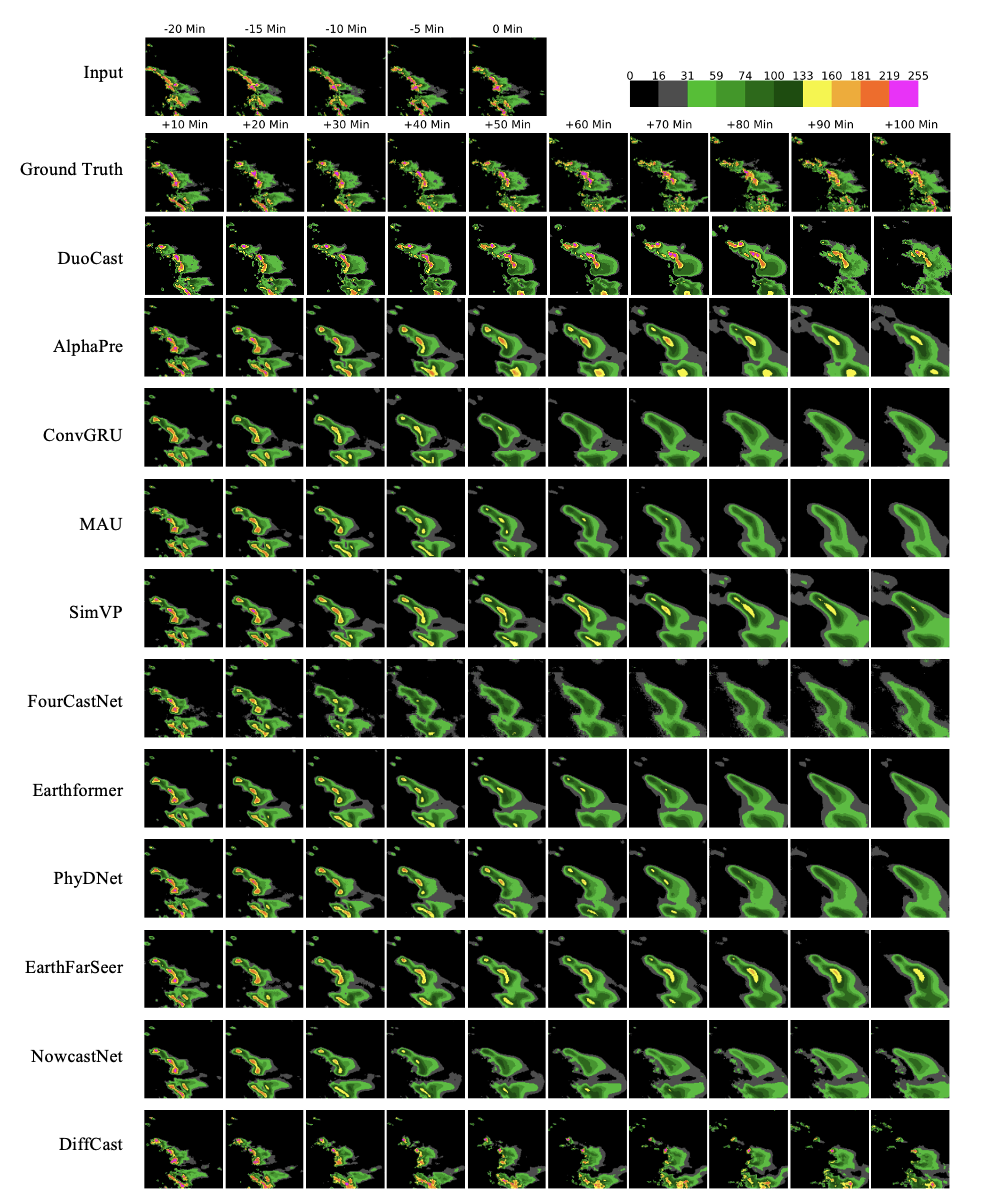}
   \caption{Prediction examples on the SEVIR dataset.}
   \label{fig:all_sevir}
\end{figure*}

\begin{figure*}[htbp]
  \centering
   \includegraphics[width=\linewidth,keepaspectratio]{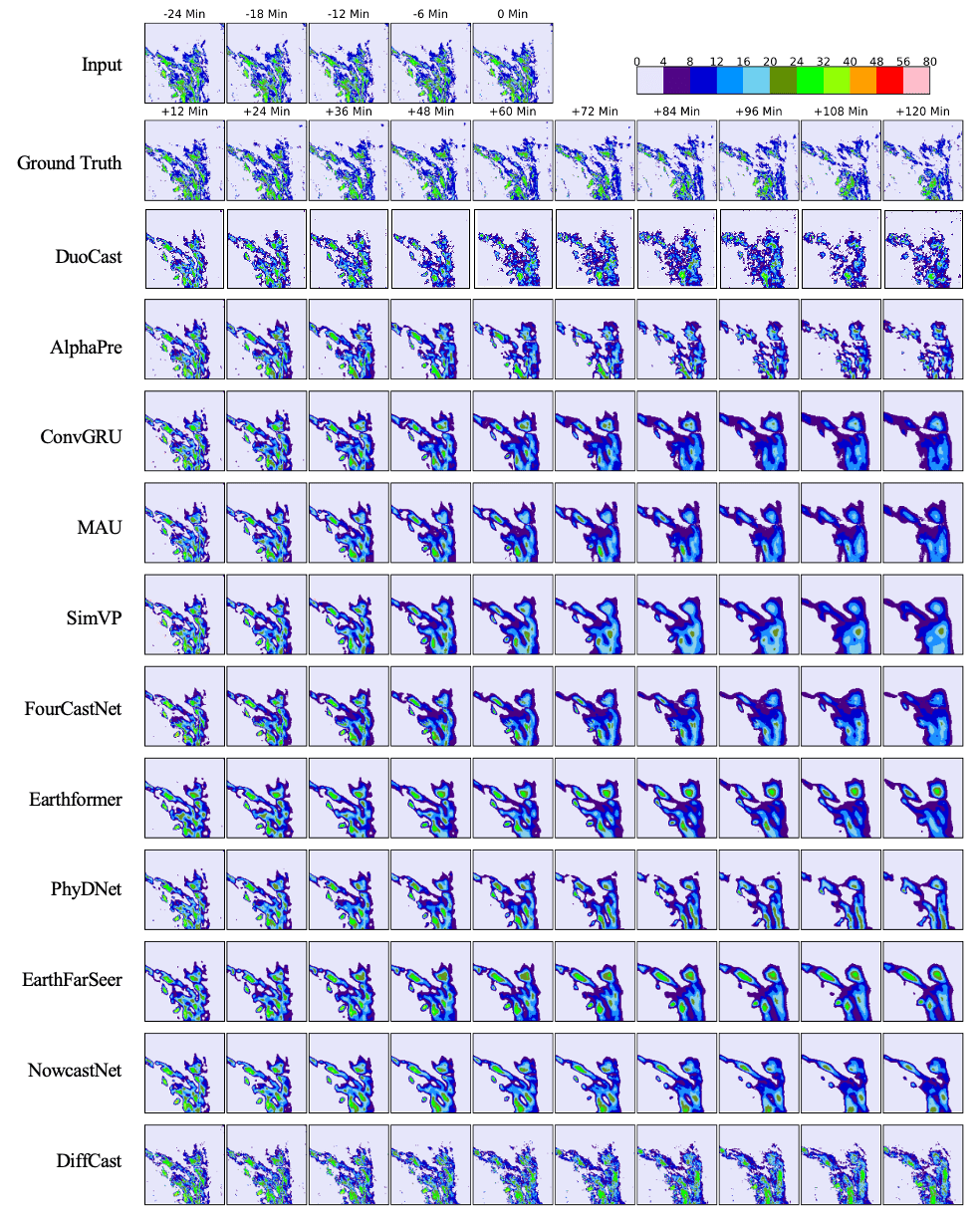}
   \caption{Prediction examples on the MeteoNet dataset.}
   \label{fig:all_meteo}
\end{figure*}

\begin{figure*}[htbp]
  \centering
   \includegraphics[width=\linewidth,keepaspectratio]{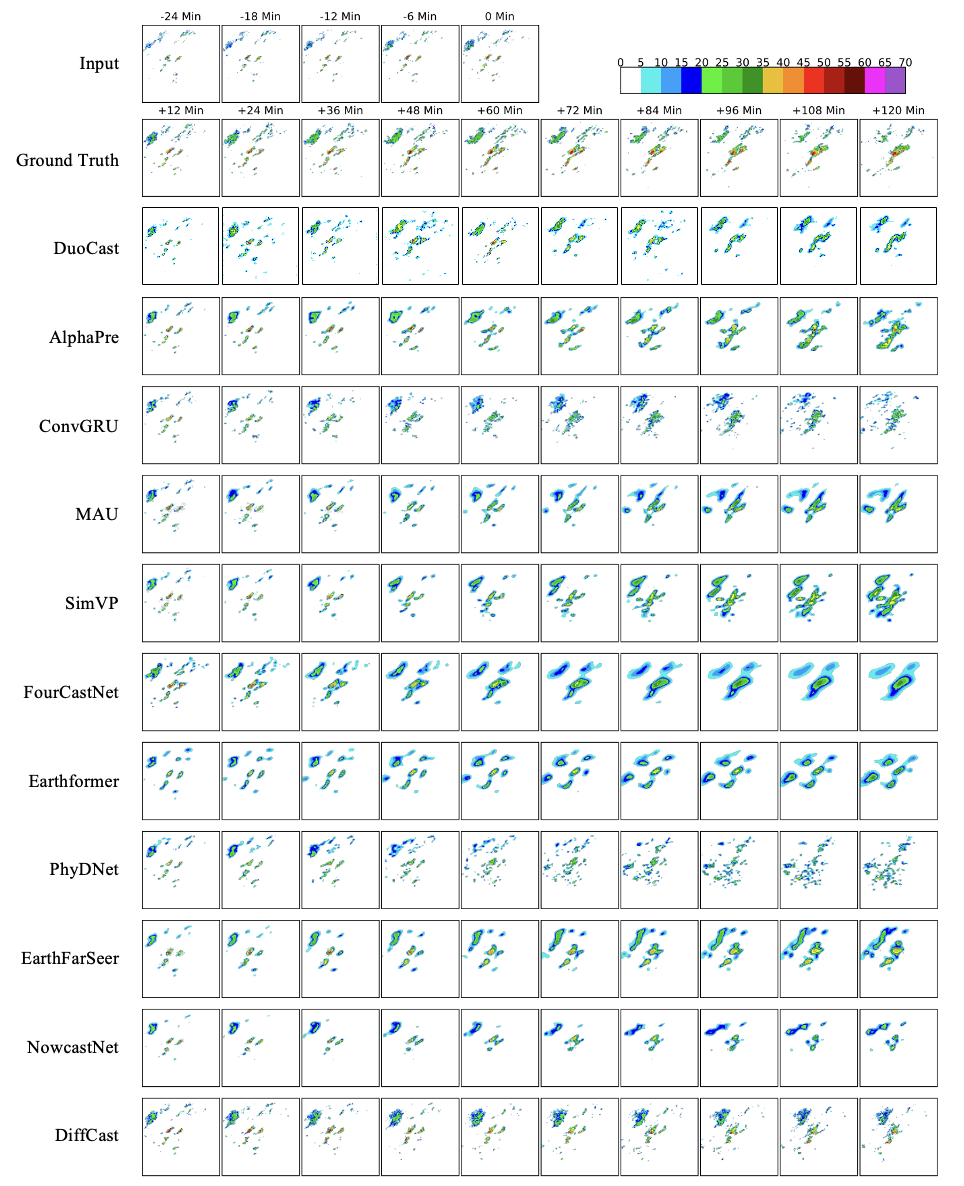}
   \caption{Prediction examples on the Shanghai dataset.}
   \label{fig:all_shanghai}
\end{figure*}

\begin{figure*}[htbp]
  \centering
   \includegraphics[width=\linewidth,keepaspectratio]{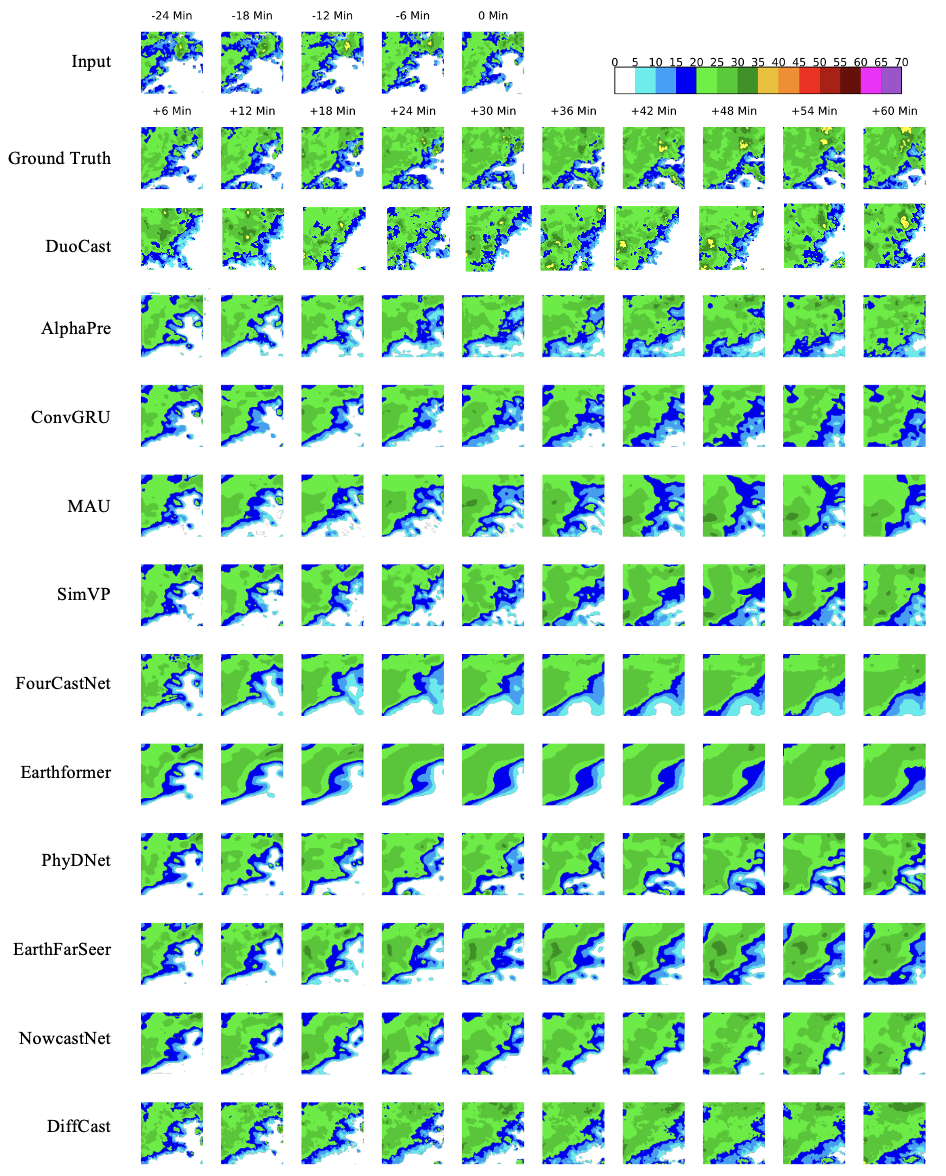}
   \caption{Prediction examples on the CIKM dataset.}
   \label{fig:all_cikm}
\end{figure*}

Additional qualitative results are shown in Figures~\ref{fig:all_sevir}–\ref{fig:all_cikm} for the four datasets. 
Compared to the most recent method AlphaPre, DuoCast generate sharper predictions and more accurately captures high-intensity precipitation regions, particularly in long lead-time forecasts. 
When compared to the baseline model DiffCast, both models perform similarly well in early frames, effectively capturing overall trends and fine-grained details. 
However, DiffCast’s predictions degrade in later frames, exhibiting spatial distortions and loss of structural consistency, whereas DuoCast preserves both detail and stability over extended time horizons.

\end{document}